# Towards integration of Privacy Enhancing Technologies in Explainable Artificial Intelligence

Sonal Allana, Rozita Dara[*], Xiaodong Lin, Pulei Xiong

School of Computer Science, University of Guelph, 474 Gordon St., Guelph, N1G 1Y4, Ontario, Canada.

[*]Corresponding author. E-mail: drozita@uoguelph.ca
Contributing authors: sallana@ uoguelph.ca; xlin08@uoguelph.ca; pulei.xiong@nrc-cnrc.gc.ca

**Abstract**

Explainable Artificial Intelligence (XAI) is a crucial pathway in mitigating the risk of non-transparency in the decision-making process of black-box Artificial Intelligence (AI) systems. However, despite the benefits, XAI methods are found to leak the privacy of individuals whose data is used in training or querying the models. Researchers have demonstrated privacy attacks that exploit explanations to infer sensitive personal information of individuals. Currently there is a lack of defenses against known privacy attacks targeting explanations when vulnerable XAI are used in production and machine learning as a service system. To address this gap, in this article, we explore Privacy Enhancing Technologies (PETs) as a defense mechanism against attribute inference on explanations provided by feature-based XAI methods. We empirically evaluate 3 types of PETs, namely synthetic training data, differentially private training and noise addition, on two categories of feature-based XAI. Our evaluation determines different responses from the mitigation methods and side-effects of PETs on other system properties such as utility and performance. In the best case, PETs integration in explanations reduced the risk of the attack by 49.47%, while maintaining model utility and explanation quality. Through our evaluation, we identify strategies for using PETs in XAI for maximizing benefits and minimizing the success of this privacy attack on sensitive personal information.

## 1. Introduction

AI systems have seen rapid adoption across a wide range of industries in recent years. The lack of transparency in black-box models is a major bottleneck in their use in high-risk domains such as medicine, defense, judiciary, among others [1], [2]. To mitigate the risk of non-transparency and to increase trustworthiness of these systems, the field of XAI has seen significant attention in recent years and researchers have proposed methods that add different degrees of transparency to black-box AI systems. However, a conflict is also found between providing explainability and protecting the privacy of individuals whose data is used in training or querying the model. Recent research has shown that explanations provide additional information about the system that can be misused by adversaries. Using explanations, adversaries can launch membership inference attacks to differentiate between members and non-members of the training set [3], [4], [5], [6], [7]. Explanations can be exploited to launch model inversion attacks to reconstruct missing information about data points [8], [9], [10], [11]. Certain types of explanations are susceptible to extraction of the model enabling adversaries to build surrogates [12], [13], [14], [15], [16] thus threatening the intellectual property [17], [18] of the model owner. The surrogate models can also become a starting point for other privacy attacks [13].

Attribute inference is a type of model inversion that infers missing input features, generally those sensitive to individuals, from outputs [19], [20]. When explanations are provided by an AI system, the attack surface of the system expands so that explanations can be alternatively exploited to launch this attack, in the place of model predictions or confidence scores [8]. Recent works [8], [9], [10] have demonstrated attribute inference on feature-based XAI, that are commonly deployed in production and machine learning as a service (MLaaS) systems to provide insights into the decision-making process of black-box models. To the advantage of adversaries, this category of XAI is found to provide a strong attack surface [8], amplifying the severity of the attack. The attack also adversely affects active XAI end-users [10] rather than a limited group of members of the training set, as in membership inference attack. Hence protection from attribute inference is essential to prevent exposure of sensitive personal information through XAI interfaces in Trustworthy AI.

The conflict of explainability with privacy [3], [21] has led to researchers proposing the use of various PETs to stem privacy leakage in XAI. Previous works have implemented differential privacy [6], [22], [23], anonymization [11], [24], [25], [26], [27] [28], [29], perturbation [30] and federated learning with cryptographic protocols [31], [32] to create privacy preserving explanations. These works utilize specific PETs mechanisms to strengthen the privacy of interpretable, example-based or feature-based explanations. However, there are only few works [6], [23], [30], [32] that evaluate the impact of PETs using known privacy attacks in XAI or attempt to create defenses for these attacks. The introduction of privacy preservation methods is also found to adversely impact other system properties such as model utility [33], [34] and explanation quality [6], [35]. Hence privacy analysis of PETs mechanisms in XAI should also consider their overall effects on the entire system.

There is currently lack of research in defenses for known privacy attacks in XAI and the effects of usage of PETs on other system properties in XAI. This has resulted in limited understanding of holistic defense techniques for specific types of attacks. Hence in this study, we address this gap by conducting a comparative analysis of PETs integration in XAI for a known privacy attack [8] and identify defense mechanisms that consider privacy with other system properties. We focus our attention on feature-based XAI due to their common usage in cloud MLaaS platforms and identify baselines and metrics for the triad of privacy, utility and explainability. We evaluate PETs at three different modelling stages of the AI lifecycle and determine strategies to maximize the benefits of PETs integration. To the best of our knowledge, this is the first comparative study on PETs integration in feature-based XAI. Our main objectives through this work are as follows:

- *Empirical evaluation of PETs in feature-based XAI:* We select commonly used PETs that are relevant to XAI from privacy preserving machine learning (PPML). Through empirical evaluation of backpropagation and perturbation-based XAI, we vary parameters and methods within a PETs category to determine strategies for privacy preservation.
- *Determination of suitable modelling stages for privacy preservation:* We integrate privacy preservation at three different AI modelling stages to determine stages and methods that work better than others. We conduct an inter-stage and intra-stage comparison to find strategies that maximize privacy preservation.
- *Determination of strategies that maximize utility:* To prevent unwanted adverse effects of PETs on the system's utility, we evaluate the impact on explanation quality, model accuracy and performance time. We seek to achieve a balance of three important properties in XAI, namely, privacy, explainability and utility.

The rest of this article is organized as follows. Section 2 presents the background of our work. We detail the methods used in the study in Section 3 and in Section 4 we describe our experimental settings. Section 5 elaborates the observations from PETs integration and the limitations of this study. In Section 6, we discuss the main results and include recommendations for PETs integration. We conclude the article in Section 7.

## 2. Background

In this section, we present background information to provide context to the work done in this study. We explain the modelling stages in AI and the main categories of feature-based XAI. Then we elaborate on the types of privacy attacks in XAI and explain the attack model for attribute inference. We summarize the existing defenses for attribute inference proposed in XAI and machine learning (ML) systems and elaborate on PETs applicable in this work. We conclude this section with the evaluation metrics used in our study.

### 2.1. Modelling stages in AI systems

Arrieta et al. [37] describe 3 stages in AI systems, namely, *pre-processing* when suitable techniques are applied before the training of the ML model, *in-processing* when techniques are applied during the training of the ML model and *post-modelling* when techniques are applied after the ML model is trained. The authors describes these modelling stages in the context of bias mitigation in AI systems. Rajabi and Etminani [38] describe pre-modelling, in-modelling and post-modelling explainability in the context of usage of knowledge graphs for generating explanations. The authors refer to pre-modelling as techniques applied on the data, in-modelling as techniques based on the inner workings of the model and post-modelling as techniques applied after training of the model. We adapt this terminology of the AI pipeline in this study, for integration of PETs before, during and after training of the model. We refer to integration of PETs on the data, prior to training of the model, as *pre-model,* the application of PETs into the training process of the model as *in-model* and that in the output of the trained model as *post-model*.

### 2.2. Feature-based XAI

XAI methods generate explanations using different approaches. Feature-based XAI is a category of methods that score input features and use the scores as explanations [37], [39]. These methods belong to two different sub-categories, namely, backpropagation and perturbation-based. Backpropagation-based use forward or backward passes to generate attributions [40] while perturbation-based add, remove, or update features and determine the change in the output [40]. In this study, we evaluate 2 methods from each sub-category, namely, integrated gradients (IG) and SmoothGrad (SG) from backpropagation-based, and Shapley Additive Explanations (SHAP) and Local Interpretable Model-agnostic Explanations (LIME) from perturbation-based. We briefly describe the main differences between these methods below:

- **IG:** This explanation method calculates the attribution of inputs by cumulating the gradients along a linear path from a baseline to the input [40]. Generally inputs where predictions are neutral are selected as baselines, such as a zero vector for text models or a black image for image models [41].
- **SG:** This method is used to sharpen gradient based saliency maps since those based on raw gradients are noisy [42]. It can be used in conjunction with other visual methods such as IG.

- **SHAP:** This method uses game theory to determine the contribution of individual features to the output [43] where features are considered as players and output is considered as pay-out [44].
- **LIME:** This method generates a local interpretable model to approximate a complex machine learning model by creating perturbations around the point of interest [45].

**2.3. Privacy risks in XAI**

Previous studies have identified three different types of privacy attacks in XAI that risk personal information of individuals or the confidentiality of the model. Table 1 summarizes the studies proposing these attacks using explanation interfaces. We briefly describe the main mechanism of the attacks below:

- **Membership inference:** This attack distinguishes members of the training set from non-members [46]. Model outputs such as predictions and confidence scores are typically exploited for this attack. However, when explanations are available, they can be misused to infer membership instead of relying on other model outputs.
- **Model inversion:** This attack uses model outputs for reconstruction or inference of missing information in query datapoints [47]. Attribute inference is a type of reconstruction that infers attributes, usually those that are sensitive, from outputs [19], [20]. When the target model provides explanations, they can be exploited to launch this attack instead of depending on other model outputs.
- **Model extraction:** This attack builds a surrogate of the victim model using an available unlabeled dataset [15] or using a data-free approach [16]. When a target model provides explanations, they can be exploited to launch this attack without the need of other model outputs.

**Table 1**

Privacy attacks in XAI.

| Privacy attack using explanations | Works |
| --- | --- |
| Membership inference | [3], [4], [5], [6], [7], [21], [23], [48] |
| Model inversion | [3], [8], [9], [10], [11], [21], [49], [50] |
| Model extraction | [7], [12], [13], [14], [15], [16], [50], [51] |

**2.4. Attribute inference attack using explanations**

In this subsection, we describe the selected privacy attack used for empirical evaluation of PETs in this work. The use of privacy attacks for empirical auditing is an emerging approach for privacy evaluation [33], [53], [54]. The selected privacy attack is proposed by Duddu and Boutet [8] and independent of the target model architecture. It is applicable when an adversary possesses an auxiliary dataset and black box access to a target model and intends to train an attack model to invert model explanations for inferring sensitive attributes. There are no defenses proposed so far in current literature for this attack on explanations.

In a typical attribute inference, an adversary can misuse white-box or black-box access to a trained model provided by a model owner. Factors such as correlation between sensitive attributes and target variable [36], [55], [56], [57], sensitive attributes' influence on the model output [19], [20] and overfitting [20] are found to contribute to the success of this attack. Models that have higher predictive power are also found to be more vulnerable to these attacks [56].

The demonstrated attack uses a multilayer neural network as the target model and a multilayer perceptron classifier as the attack model. Fig. 1, adapted from [8], demonstrates one of the two threat models described in the work, where *s* indicates the sensitive attribute to be inferred and *x* indicates the non-sensitive attributes in the input. When presented with an input ($x \cup s$), which comprises of both sensitive and non-sensitive attributes, the target model generates two types of outputs, namely, predictions $f_{target}(x \cup s)$ and explanations $\phi(x \cup s)$. The adversary can use both or either output for training the attack model for inferring the sensitive attribute *s*. The threat model assumes no access to the input interface by the adversary, though predictions and explanations are accessible for arbitrary inputs. The threat model also assumes that the attacker is aware of the statistical distribution of the training dataset. This is achieved by the authors by partitioning the original dataset and using a portion for training and another for testing the attack model.

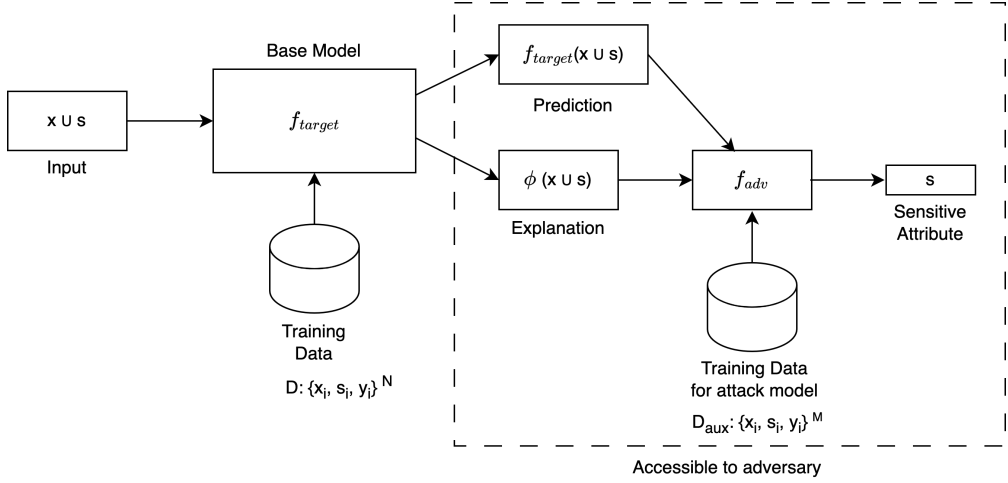

Fig. 1. Threat model of attribute inference using explanations.

The study [8] determined that model explanations provide a strong attack surface and sensitive attributes can be inferred from explanations even when predictions are excluded. Thus, the adversary can infer *s* from the explanations function, $\phi$, by exploiting the following mapping of the attack model:

$$f_{adv}: \phi(x \cup s) \to s \qquad (1)$$

The study also demonstrated that the attack is effective even when sensitive attributes are excluded from the training data.

## 2.5. Defenses against attribute inference attacks

Researchers have attempted different strategies for mitigation of attribute inference through predictions and confidence scores. However, there is limited work on strategies for defending the attack through explanations. Currently there are few works that explore the mitigation of this attack in XAI. For explanations generated by feature-based and interpretable models, C. L. Chen et al. [36] described a method of limiting inference of sensitive attributes by defining a maximum confidence measure and perturbing the decision mapping of an algorithm. The method enabled the release of transparency reports in a privacy preserving manner. In a different study on an example-based explanation type, namely, counterfactuals, Molhoek and Laanen [32] built secure explanations from vertically partitioned data in a two-party setting. The setup was found to be resilient to attribute inference. Wu et al. [31] designed privacy preserving decision trees in vertical federated learning and enhanced the training protocol to achieve resilience to attribute inferene. The work employed threshold partially homomorphic encryption and additive secret sharing for intermediate exchanges of messages between participating clients.

Other non-XAI works have proposed perturbation techniques to mitigate attribute inference by adding noise to confidence scores [58] or publicly available information [59]. Strategies such as rounding the confidence scores [47], privacy-guided training to reduce a model's reliance on sensitive features [19] and use of synthetic data to replace the original training dataset [55] have also been proposed. The main consideration of these countermeasures is to achieve privacy of sensitive attributes while reducing any adverse effects on the system utility.

### 2.6. PETs integration in XAI

PETs are the privacy preservation methods employed to safeguard data [60] in computer systems. There are different categories of PETs available in PPML such as anonymization, cryptography, differential privacy, perturbation techniques and federated learning [61], [62], [63], [64]. We scope our study to use one PETs category in each AI modelling stage while using distinct categories across the three stages. We exclude federated learning due to use of centralized learning in our study. In pre-model, a suitable PETs is applied on the data, prior to training of the model. We select anonymization for this stage due to its simplicity in comparison to cryptographic methods. In in-model, PETs is applied on the training process of the model. We select differential privacy due to its previous usage [6], [23] in this stage for mitigation of membership inference in XAI. In post-model, PETs is applied after the model is trained. We select perturbation techniques on the model outputs due to its previous usage [30] in mitigation of data reconstruction in XAI. Within each category, we choose a method unevaluated on the selected attribute inference attack on explanations. Thus, we select synthetic data, DP-SGD training algorithm and noise from anonymization, differential privacy, and perturbation techniques respectively. We perform an ablation study by introducing one privacy preservation method at a time in the pipeline to enable inter-stage comparison. Varying algorithms or parameters in each stage, allows intra-stage comparison. Below we elaborate the PETs selected for this study and their previous application in privacy preservation in XAI.

### 2.6.1. Synthetic data

Synthetic data are surrogate datasets [60] that have similar distribution as the original dataset. Those generated by statistical or ML methods, capture important patterns from the original data while lacking one-on-one mapping [65]. Due to the inherent privacy preserving characteristics of synthetic data, their use for anonymization of sensitive training data has gained traction in PPML. The technique when applied on private datasets can produce non-privatized versions suitable for training.

The use of synthetic images for privacy preserved explanations was proposed by Campos et al. [26] to anonymize the training data in medical domain. Latent diffusion models were used for generation of the synthetic images and those similar to the training data were removed to protect privacy of sensitive information. Abbasi et al. [66] generated synthetic datasets using differential privacy for training different types of models. The authors used a similarity score to determine the change in explanations due to privacy preservation, and utility loss to measure the drop in model accuracy.

We use an approach similar to [66] and apply this PETs to replace sensitive training datasets with their non-private surrogates. We utilize the following synthetic data generation algorithms based on statistical and ML methods:

- Conditional Generative Adversarial Network (CTGAN) [67]: This algorithm uses a conditional generator on tabular data to synthesize both continuous and discrete values.
- Gaussian Copula [68]: This algorithm uses a classic statistical technique for generation of tabular data.
- Tabular Variational Encoder (TVAE) [67]: This algorithm adapts a variational autoencoder for generation of tabular data.

### 2.6.2. Differential Privacy (DP)

DP is a well-known technique with a mathematical guarantee of privacy. It enables to mask a record, such that the outcomes from two adjacent datasets, that differ by a single record, is indistinguishable [69]. The parameter, $\varepsilon$, also known as privacy budget, measures the amount of privacy that can be achieved. Lower values of $\varepsilon$ indicate higher privacy and vice versa. Mathematically, if $x$ and $y$ are two adjacent datasets that differ by a single record and $M$ is a property or mechanism on the datasets, for all subsets $S \subseteq Range(M)$ and $\varepsilon > 0$, M gives $\varepsilon$-DP if the following is satisfied [69]:

$$Pr\,[M(x) \in S] \leq exp(\varepsilon) \times Pr\,[M(y) \in S] \qquad (2)$$

A mechanism is said to give $(\varepsilon, \delta)$-DP if the following relation holds [70]:

$$Pr\,[M(x) \in S] \leq exp(\varepsilon) \times Pr\,[M(y) \in S] + \delta \qquad (3)$$

In neural networks, $(\varepsilon, \delta)$-DP can be achieved using differentially private stochastic gradient descent (DP-SGD) [71] optimization algorithm. It involves computing the gradients for a randomly selected subset of samples, clipping the gradients at a set threshold, and adding noise to the average. This results in reducing the influence of any specific example [71] in the outcome of the model and thus privatizes the training process. Liu et al. [6] used this approach in feature-based XAI for privatizing the target model and evaluating its effect on membership inference. The work found mitigation of the attack at high privacy budgets resulting in degradation of model utility. Similar observations were noted by Cohen & Giryes [23] on applying DP to the training process for explanations using self-influence functions. However, both these works evaluate membership inference and the effect of DP training on attribute inference from explanations is not yet known.

### 2.6.3. Noise perturbations

The use of noise for protection of sensitive data is a classical technique used with statistical databases. Noise techniques typically involve obfuscating data, such as attribute or class values [72], to preserve privacy. Various methods of noise addition are described in literature such as the use of additive [73], multiplicative [74] or calibrated [75] noise, with the privacy guarantee depending on the technique used for perturbations [60].

Jeong et al. [30] applied noise to image explanations to mitigate model inversion and prevent recovery of query images in facial recognition models. The work used a generative network for noise generation and an inversion network to invert the explanations. The method was found to reduce the leakage from explanations by introduction of infinitesimal noise and without the need of retraining of models. We use a similar approach of adding noise to explanations from the trained model. However, instead of recovering input images from explanations, the attack we employ recovers sensitive features in inputs. We use two types of calibrated noise, namely, random and differentially private, with Laplace and Gaussian distributions.

### 2.7. Evaluation Metrics

We evaluate 3 properties of XAI systems in this study, namely, privacy, explainability and utility. For measuring privacy, we launch the selected attribute inference attack on the target model and use attack effectiveness as a measure of privacy leakage. For explainability, we use XAI faithfulness metrics to determine the quality of explanations. We measure utility using the model accuracy and performance time. Thus, we base our evaluation on the 3 groups of evaluation metrics below.

#### 2.7.1. Privacy

These metrics are applicable to the attack model and enable to gauge the privacy leakage by measuring the effectiveness of privacy attacks. We have identified the following applicable metrics:

- **Attack Precision (AP):** This is the ratio of true positives (TP) and sum of true positives and false positives (FP) detected by the attack model, given by:

$$AP = \frac{TP}{TP + FP} \qquad (4)$$

- **Attack Recall (AR):** This is the ratio of true positives and sum of true positives and false negatives (FN) detected by the attack model, given by:

$$AR = \frac{TP}{TP + FN} \qquad (5)$$

- **Attack F1-Score:** This is the harmonic mean of the attack precision and the attack recall given by,

$$Attack\ F1 - Score = \frac{2 \times AP \times AR}{AP + AR} \qquad (6)$$

- **Attack Success:** This is the ratio of the number of successful attacks to the total number of attacks [76], given by,

$$Attack\ Success = \frac{\#Successful\ attacks}{\#All\ attacks} \qquad (7)$$

By definition, attack F1-score is a representative measure for both attack precision and recall. A successful attack is indicated by high attack F1-scores and/or attack success. A successful attack mitigation method is expected to lower the attack F1-scores and/or attack success in comparison to a baseline.

#### 2.7.2. Explainability

XAI metrics enable evaluation of explanations on different criteria such as faithfulness, robustness, localization, complexity, randomization and axiomatic [77]. In this study, we will focus on faithfulness metrics that indicate the alignment of explanations with the predictive behaviour of the model [77]. We have identified the following applicable metrics:

- **Faithfulness Correlation:** Faithfulness correlation measures the correlation between the sum of attributions in a subset of features and the difference in output when those features are set to a baseline value [78].

- **Faithfulness Estimate:** Faithfulness estimate computes the aggregate value of the correlation between probability drops in the predicted class when certain features are obscured or removed, and the relevance scores on various points [79].
- **Sufficiency:** This property indicates the extent to which, if two datapoints hold the same explanation, then they can be expected to have the same prediction outcomes [80].

An increase in the values of the above metrics is considered as an improvement of explanation quality.

### 2.7.3. Utility

The following metrics are applicable to the target model and useful for evaluating its utility:

- **Accuracy:** This is the number of correctly classified samples to the total number of samples. A high accuracy is desired for high model utility.
- **Training Time:** This is the time needed to train the model for the given number of epochs. A low training time is preferable for fast development and deployment of AI models.

## 3. Method

In this study, we apply PETs to create defense mechanisms in feature-based XAI against attribute inference attack. We consider both sub-categories of feature-based XAI, namely, backpropagation and perturbation-based. Within each sub-category, we select two XAI algorithms, namely, IG and SG from backpropagation-based, and SHAP and LIME from perturbation-based methods. We use a non-private neural network model [33] as the baseline. The attack methodology is independent of the model architecture and is applicable to non-neural network models as well. The target model is trained on datasets that contain sensitive features such as age, race and gender. The selected XAI algorithms are applied on the non-private model to generate explanations. The attribute inference attack is applied on the explanations from the baseline model. We apply privacy, explainability and utility metrics on this system to get the baseline measurements that can be used for comparison with PETs enhanced systems.

We apply PETs at a time, at three stages of the AI pipeline. In pre-model, PETs are applied on the training data in the form of synthetic data; in in-model, PETs are applied during the training process in the form of DP-SGD optimization algorithm and in post-model, PETs are applied in the form of noise on the model explanations. In each setup, we generate explanations using the selected XAI methods and launch the attribute inference attack on private explanations. We conduct an inter-stage comparison of attack effectiveness to determine strategies of mitigation. An intra-stage comparison enables to find better approaches and tuning strategies within a stage. XAI faithfulness metrics enable to determine any undesirable effects on explanations. Model utility metrics determine the effect of privacy preservation on the utility of the model. Below we elaborate the setup of the baseline system followed by the setup used in each of the three stages.

### 3.1. Baseline Setup

We use the non-private model described in [33] to build the baseline setup. A multilayer neural network for binary classification problems is used as the target model and trained on selected state-of-the-art datasets. All input features in the datasets, including sensitive attributes, are used for training. Appropriate tuning of hyperparameters is used to achieve state-of-the-art accuracy on each dataset. Applicable metrics for explainability and utility are applied to give the baseline measurements.

To evaluate the privacy of the system, the selected attribute inference attack is launched on each baseline explanation and privacy metrics are applied to determine its effectiveness. Each attack attempts to infer two binary sensitive attributes from the input features. The attack model infers sensitive information using only explanations and excludes predictions. The setup of the baseline system in Fig. 2 differs from Fig. 1 in that we eliminate predictions from training the attack model and use only explanations for inferring sensitive attributes at runtime. This is feasible since explanations alone are shown to provide a strong attack surface [8].

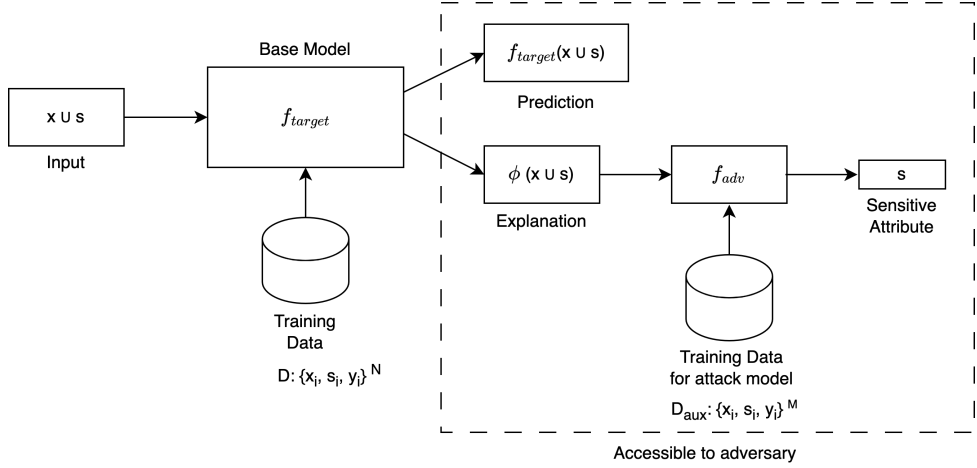

*Fig. 2. Attribute inference attack using only explanations.*

### 3.2. Pre-model Setup

PETs are applied to the training data to create surrogate private datasets which can be used for training in place of the original datasets. We generate data that is realistic using 3 different algorithms, namely, CTGAN, Gaussian Copula and TVAE. The setup of the privacy enhanced system in this stage is as shown in Fig. 3. Thus, the target model is trained on synthetic data instead of the real data. The attribute inference attack is applied to explanations from this privatized system and the identified metrics are applied to determine the effect on privacy, explainability and utility.

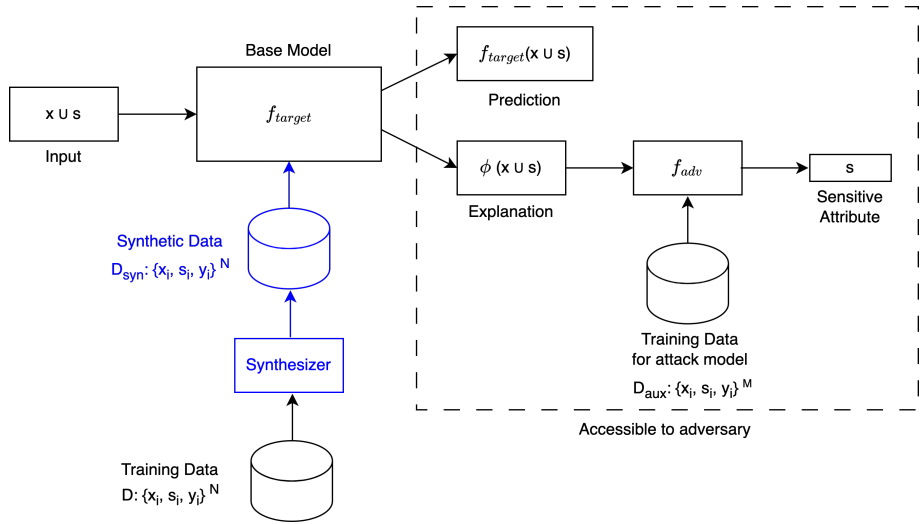

*Fig. 3. Pre-model setup with attribute inference attack.*

## 3.3. In-model Setup

PETs are applied to the model during its training process in the form of DP-SGD algorithm. This gives the privatized predictions and explanations as shown in Fig. 4. Models with different values of privacy budget, $\varepsilon$, are built in this stage. We consider two tiers of privacy guarantees, namely, strong formal ($\varepsilon \leq 1$) and reasonable ($\varepsilon \leq 10$) [54] and choose values of $\varepsilon$ distributed over this range. The attribute inference attack is repeated on each DP model and metrics are applied for privacy, explainability and utility. The impact of varying $\varepsilon$ across the two tiers of privacy protection on the effectiveness of the attack is observed.

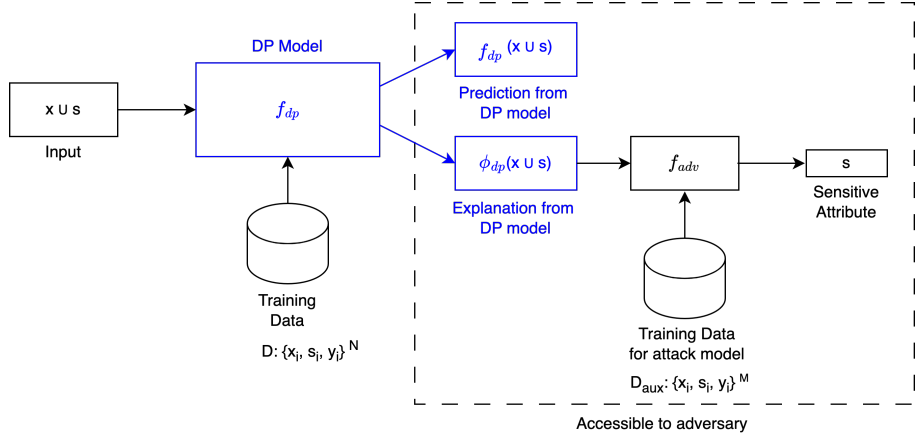

*Fig. 4. In-model setup with attribute inference attack.*

## 3.4. Post-model Setup

During this stage, the target model is unchanged. PETs are applied to the model in the form of noise addition at the explanation interface, altering only explanations and leaving the predictions unchanged as shown in Fig. 5. We use two types of noise, namely, random and calibrated noise. We use random noise with Laplace and Gaussian distribution where the respective scale and standard deviation is randomly selected. We calibrate noise using DP with Laplace or Gaussian mechanisms. The value of $\varepsilon$ is set to 1 after observing $\varepsilon$ across a range of values in the strong formal tier ($\varepsilon \leq 1$) [54] and determining no substantial change in privacy by varying its value. Each explanation is perturbed with noise addition, to produce four variations of explanations, namely, random Laplace, random Gaussian, DP Laplace, and DP Gaussian. The attribute inference attack is applied on each type of noisy explanation and the metrics are used on privacy, explainability and utility.

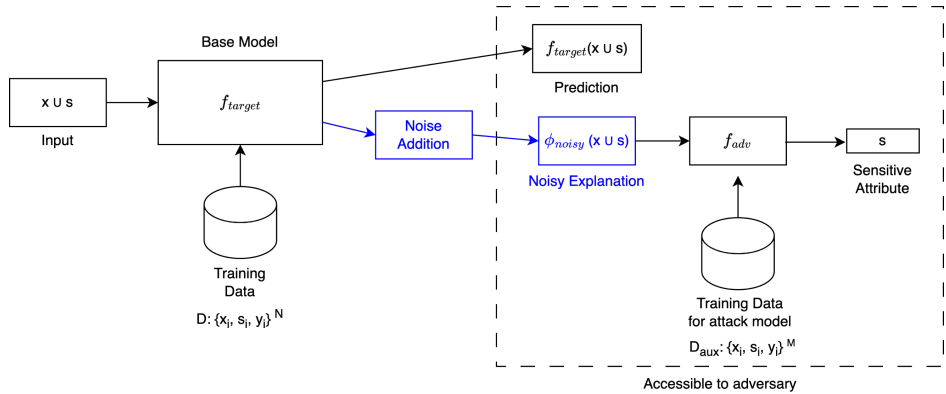

*Fig. 5. Post-model setup with attribute inference attack.*

## 4. Experimental Settings

In this section, we describe the experimental settings we employ for empirical evaluation of the baseline and PETs enhanced systems. To facilitate reproducibility of our experiments, we make our code available publicly at https://github.com/dmpglab/ppxai.

### 4.1. Datasets

We consider 4 state-of-the-art datasets used in previous works on privacy attacks using explanations [3][8]. These cover different application domains such as finance, medical and justice. All datasets are binary classification problems and tabular in structure. Table 2 highlights the main properties of the datasets. We summarize the characteristics below:

1. Adult [81]: This dataset consists of records of 48,842 individuals from 1994 US census database and indicates if individuals' annual salary exceeds 50K. It consists of 14 features. We use sex and race as sensitive attributes.
2. Credit [82]: This dataset consists of default payment records of 30,000 individuals from Taiwan. It consists of 23 features. We use sex and age as sensitive attributes.
3. Compas [83]: This dataset consists of 7,214 records of criminal defendants from Broward Country, Florida from 2013 and 2014 collected by Propublica to test a commercial recidivism risk assessment tool called COMPAS (Correctional Offender Management Profiling for Alternative Sanctions). We use a subset of 10 features to train the model out of 53 in the original dataset. We use sex and race as sensitive attributes.
4. Hospital [84]: This is a modified version of clinical care records of diabetic patients of 130 hospitals and integrated delivery networks in the United States from 1999-2008 extracted from Health Facts database [85]. It consists of 101,766 records with 127 features out of which 10,000 records for each class are sampled. It consists of sensitive fields such as gender and race.

**Table 2**

Dataset properties.

| Properties | Adult | Credit | Compas | Hospital |
|---|---|---|---|---|
| **Number of records** | 48,842 | 30,000 | 7,214 | 101,766 (20,000 extracted) |
| **Records after preprocessing** | 32,561 | 30,000 | 6,907 | 20,000 |
| **Number of records in positive / negative class** | 7,841 / 24,720 | 6,636 / 23,364 | 2,519 / 4,388 | 10,000 / 10,000 |
| **Number of features** | 14 | 23 | 10 | 127 |
| **Sensitive features** | Sex, Race | Sex, Age | Sex, Race | Gender, Race |
| **Binary classification problem** | Individual's salary > 50K | Individual will default payment | Individual will reoffend (recidivism) | Individual will be readmitted |

### 4.2. Preprocessing

Prior to training the models or generation of synthetic data, all 4 datasets are preprocessed for removal of missing values, one-hot encoding of categorical data and conversion of the target variable into binary. We adapted the preprocessing of Adult, Credit and Hospital from [33], [8] and [84] respectively. For Compas, we retrieved the two years scores from Propublica, dropped direct identifiers, date fields and sparse columns to get 10 features for training. We follow the procedure in [8] for selection of two sensitive fields and conversion of the values of these attributes into binary. The criteria for conversion of the sensitive fields into binary is summarized in Table 3.

**Table 3**

Selected sensitive attributes for inference.

| Properties | Adult | Credit | Compas | Hospital |
|---|---|---|---|---|
| **Sensitive attribute 1** | Sex | Sex | Sex | Gender |
| value = 1 | Male | Female | Male | Male |
| value = 0 | Not Male | Male | Not Male | Not Male |
| **Sensitive attribute 2** | Race | Age | Race | Race |
| value = 1 | White | < 40 | Caucasian | Caucasian |
| value = 0 | Not White | >= 40 | Not Caucasian | Not Caucasian |

We followed the train-test split described in [8] and used 67% of each dataset for training and 33% for testing. The test set was utilized as the auxiliary dataset and further split into 50-50% for training and testing the attack model. To understand if the success of attribute inference was due to correlation of the sensitive attributes with the target variable or with the non-sensitive attributes, we determined the Pearson coefficient (Table 4) between them. As seen, the correlation is low ($< 0.3$) between the sensitive attributes and the target variable, Y. The correlation is also low ($< 0.1$) between the sensitive attributes and the other non-sensitive attributes in the datasets. Thus, this observation precludes correlation between sensitive attributes and target variable or non-sensitive attributes from being the cause of attribute inference.

**Table 4**

Pearson correlation of sensitive attributes.

| Dataset | Y | | Non sensitive attributes | |
|---|---|---|---|---|
| | Sensitive attribute 1 | Sensitive attribute 2 | Sensitive attribute 1 | Sensitive attribute 2 |
| Adult | 0.2160 | 0.0852 | $0.0024 \pm 0.0273$ | $-0.0116 \pm 0.0129$ |
| Credit | -0.0400 | -0.0245 | $-0.0305 \pm 0.0199$ | $0.0050 \pm 0.0350$ |
| Compas | 0.0514 | -0.1705 | $0.0293 \pm 0.0396$ | $-0.0259 \pm 0.0590$ |
| Hospital | -0.0184 | 0.0185 | $0.0003 \pm 0.0018$ | $0.0017 \pm 0.0027$ |

### 4.3. System Architecture

In this subsection, we describe the architecture of the target and attack models. We also detail the libraries and tools employed for XAI and PETs integration in the different pipeline stages.

#### 4.3.1. Baseline Model

We used a multilayer neural network architecture adapted from Blanco-Justicia et al. [33] as the base classification model. It comprises of a 2-layer fully connected neural network with 40 neurons in each layer and ReLU activation for all datasets, except Hospital, which gave a low accuracy with this architecture. Hence for Hospital dataset, we adapted the architecture from Shokri et al. [3] to use a 4-layer connected neural network with 1024, 512, 256, 128 neurons respectively in each layer and applied ReLU activation to the neurons. Binary cross entropy was used as the loss function with Adam optimizer. The hyperparameters were appropriately tuned and set to 50 epochs, learning rate of 1e$^{-3}$ and batch size of 48 for achieving state-of-the-art accuracy. Each training process was repeated 5 times, and the average accuracies and training times are reported in Table 5. We achieved an average test accuracy of 84.58% on Adult, 75.07% Credit, 75.84% on Compas and 56.54% on Hospital, which are the state-of-the-art values reported in other studies [3], [8], [33]. All training was done on a 2 GHz Quad-Core CPU with 16 GB 3733 MHz RAM. We used TensorFlow Keras for building the target model architecture and for training.

**Table 5**

Average accuracies and training times of baseline models.

| Properties | Adult | Credit | Compas | Hospital |
|---|---|---|---|---|
| **Training Accuracy (%)** | 86.47 | 75.22 | 76.12 | 72.67 |
| **Testing Accuracy (%)** | 84.58 | 75.07 | 75.84 | 56.54 |
| **Training Time (secs)** | 45.05 | 39.90 | 27.76 | 247.77 |

### 4.3.2. Attack model

We used the attack model described in [8] which comprises of a 3-layer multilayer perceptron classifier with 64, 128 and 32 neurons respectively in each layer. The model was built using scikit-learn library. It uses Adam optimizer and a L2 regularization term of 1e$^{-3}$. The maximum number of epochs was set to 500. The same attack model was employed to attack the baseline and each of the three setups. Each attack was repeated 5 times, and we report the average values of the metrics. The total number of attacks on the PETs enhanced systems, are as computed in (8), (9) and (10) below:

$$Total\ number\ of\ pre-model\ attacks = 3\ synthetic\ generation\ algorithms \times 4\ explanations \times 4\ datasets \times$$
$$2\ senstitive\ attributes = 96\ attacks \qquad (8)$$

$$Total\ number\ of\ in-model\ attacks = 4\ privacy\ budgets \times 4\ explanations \times 4\ datasets \times$$
$$2\ sensititve\ attributes = 128\ attacks \qquad (9)$$

$$Total\ number\ of\ post-model\ attacks = 4\ noise\ types \times 4\ explanations \times 4\ datasets \times$$
$$2\ sensitive\ attributes = 128\ attacks \qquad (10)$$

Thus, in all, we conducted 352 attacks on the PETs enhanced systems.

### 4.3.3. XAI Methods

We imported libraries for respective explanation methods in TensorFlow. For IG and SG, we used INNvestigate [86] library. For SHAP and LIME, we used the shap [87] and lime [88] libraries respectively. We implemented faithfulness metrics for evaluation of explanations using the Quantus [77] library. All libraries are compatible with TensorFlow.

### 4.3.4. Synthetic Data Generation

We used Synthetic Data Vault (SDV) [89] for generation of synthetic data using CTGAN, Gaussian Copula and TVAE methods. After generation of the synthetic datasets, we ran diagnostic tests offered by the library for each dataset to verify the data validity and determine structural issues. As seen in Table 6, we get a perfect score of 100% as required [90] on all datasets for each synthetic generation algorithm.

**Table 6**

Diagnostic test on properties of synthetic data.

| Dataset | Synthetic Algorithm | Data Validity | Data Structure |
|---|---|---|---|
| **Adult** | CTGAN | 1.0 | 1.0 |
| | Gaussian Copula | 1.0 | 1.0 |
| | TVAE | 1.0 | 1.0 |
| **Credit** | CTGAN | 1.0 | 1.0 |
| | Gaussian Copula | 1.0 | 1.0 |
| | TVAE | 1.0 | 1.0 |
| **Compas** | CTGAN | 1.0 | 1.0 |
| | Gaussian Copula | 1.0 | 1.0 |
| | TVAE | 1.0 | 1.0 |
| **Hospital** | CTGAN | 1.0 | 1.0 |
| | Gaussian Copula | 1.0 | 1.0 |
| | TVAE | 1.0 | 1.0 |

### 4.3.5. DP Model

We applied the same architecture of the baseline model to the DP models. The differentially private counterpart of Adam optimizer from TensorFlow privacy library was applied for optimization [54]. TensorFlow Keras was used for building the DP model. Similar to the baseline model, the DP model was trained for 50 epochs, however the hyperparameters such as learning rate, L2 norm clipping threshold, batch and microbatch sizes were tuned to different datasets to maximize the utility. The final tuned hyperparameters for each dataset are outlined in Table 7.

**Table 7**

Hyperparameters for DP Training.

| Hyperparameters | Adult | Credit | Compas | Hospital |
|---|---|---|---|---|
| **Learning rate** | 15e-5 | 15e-5 | 15e-5 | 0.01 |
| **Epochs** | 50 | 50 | 50 | 50 |
| **Batch size** | 5 | 48 | 1 | 4096 |
| **Microbatch size** | 5 | 12 | 1 | 16 |
| **Delta** | 1e-6 | 1e-6 | 1e-6 | 1e-6 |
| **L2 norm clip** | 1e-5 | 1e-5 | 1e-3 | 1 |

TensorFlow privacy library was used to train the models with DP-SGD algorithm to achieve $(\varepsilon, \delta)$-DP. The hyperparameter $\delta$ was set to 1e-6 so that as recommended in literature, it is $\ll \frac{1}{number\ of\ dataset\ records}$ [91]. We used 4 different $\varepsilon$ values to target the strong formal ($\varepsilon \leq 1$) and reasonable ($\varepsilon \leq 10$) tiers of privacy guarantees. Each $\varepsilon$ value can be achieved by addition of noise referred to as noise multiplier. Due to the difficulty of calculating the noise multipliers for exact target values of $\varepsilon$ in TensorFlow privacy [33], we used approximate values close to target $\varepsilon$ values of 0.01, 0.1, 1 and 5 as indicated in Table 8. The average training and test accuracies, training time for each value of noise multiplier and privacy budget are also summarized in Table 8.

**Table 8**

Average accuracies and training times of DP models.

| Properties | Adult | | | | Credit | | | |
|---|---|---|---|---|---|---|---|---|
| | $\varepsilon \sim 0.01$ | $\varepsilon \sim 0.1$ | $\varepsilon \sim 1$ | $\varepsilon \sim 5$ | $\varepsilon \sim 0.01$ | $\varepsilon \sim 0.1$ | $\varepsilon \sim 1$ | $\varepsilon \sim 5$ |
| $\varepsilon$ | 0.01 | 0.11 | 0.97 | 5.01 | 0.01 | 0.11 | 0.97 | 5.01 |
| Noise Multiplier | 4000 | 500 | 65.84 | 14.68 | 4000 | 500 | 65.84 | 14.68 |
| Training Accuracy (%) | 74.21 | 75.84 | 75.81 | 75.85 | 38.16 | 52.2 | 62.38 | 68.60 |
| Testing Accuracy (%) | 73.65 | 76.07 | 76.14 | 76.05 | 37.87 | 50.73 | 63.38 | 69.33 |
| Training Time (s) | 489.87 | 421.77 | 517.06 | 481.30 | 63.17 | 64.33 | 66.77 | 69.38 |

| Properties | Compas | | | | Hospital | | | |
|---|---|---|---|---|---|---|---|---|
| | $\varepsilon \sim 0.01$ | $\varepsilon \sim 0.1$ | $\varepsilon \sim 1$ | $\varepsilon \sim 5$ | $\varepsilon \sim 0.01$ | $\varepsilon \sim 0.1$ | $\varepsilon \sim 1$ | $\varepsilon \sim 5$ |
| $\varepsilon$ | 0.01 | 0.11 | 0.97 | 5.01 | 0.01 | 0.11 | 0.97 | 5.01 |
| Noise Multiplier | 4000 | 500 | 65.84 | 14.68 | 4000 | 500 | 65.84 | 14.68 |
| Training Accuracy (%) | 63.28 | 63.57 | 63.48 | 63.83 | 49.97 | 50.22 | 50.30 | 50.97 |
| Testing Accuracy (%) | 65.70 | 63.70 | 63.34 | 63.49 | 49.95 | 50.48 | 50.19 | 50.48 |
| Training Time (s) | 567.75 | 389.05 | 388.74 | 367.36 | 214.09 | 220.07 | 218.64 | 217.96 |

### 4.3.6. Post Noise Addition

For adding random noise, we randomly generated the scale and standard deviation of Laplace and Gaussian noise respectively. We used Google's DP library [92] for generating noise calibrated using DP. We utilized the Java API of this library to add noise asynchronously to each explanation value. The noise was added sequentially to the explanation records once they were generated from the target models.

## 5. Experimental Results

In this section, we report the privacy, explainability and utility results from the PETs enhanced systems. Each sub-section focusses on the results observed on a specific measured property from different setups in the AI pipeline.

## 5.1. Impact of PETs on privacy

We observe the attack effectiveness on the baseline system and on each PETs enhanced system using attack success and attack F1-scores on two sensitive attributes for each dataset. Fig. 6 displays the results on the baseline systems for each explanation. For sensitive attribute 1, the attack success readings in Fig. 6 (a) and the attack F1-score readings in Fig. 6 (c) are aligned. For sensitive attribute 2, similar observations are seen from the attack success in Fig. 6 (b) and attack F1-score in Fig. 6 (d). Thus, these two metrics are observed to be aligned and indicate similar trends in the attack effectiveness. We also observe that among backpropagation-based methods, SG has lower attack effectiveness on all datasets compared to IG. Among perturbation-based methods, LIME is seen to be more resilient compared to SHAP. Overall, Credit dataset is observed to be resilient to the attack in comparison to Adult, Compas and Hospital. The attack success and attack F1-scores are lowest on this dataset and nearest to random guess (red dotted line in Fig. 6 (a) and 6 (b)) in comparison to other datasets.

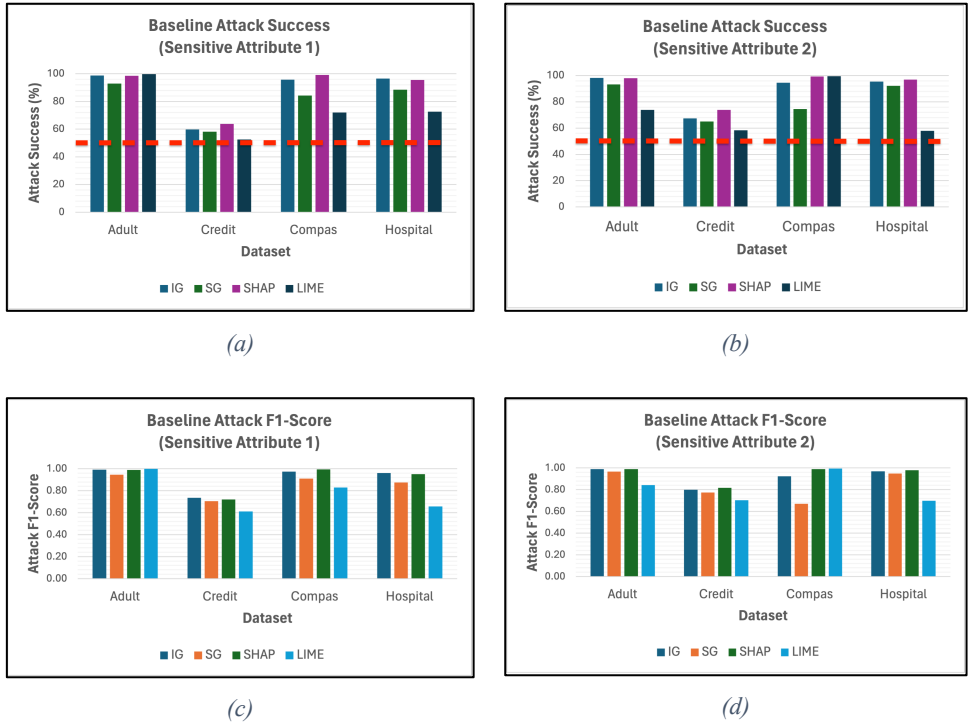

*Fig. 6. Baseline attack effectiveness for all explanation types.*

Next, we visualize the results of attack effectiveness in pre-model, in-model and post-model setups in Fig. 7 and 8 for each sensitive attribute respectively. Similar to the observation on the baseline systems, we find that the attack success and attack F1-scores are aligned on all explanations across the 3 setups. Due to the similarity in the observations of these two metrics, we discuss the attack success in this subsection and include the results from attack F1-scores in the appendix. In each sub-figure in Figs. 7 and 8, the baseline attack success is reported alongside other algorithms or tuned hyperparameters within a PETs category. Fig. 7 displays the attack success in PETs enhanced models in pre-model, in-model and post-model setups for sensitive attribute 1 while Fig. 8 displays the same for sensitive attribute 2. The horizontal red dotted line in each sub-figure indicates random guess.

The results from pre-model indicate that synthetic data is ineffective at mitigating attribute inference. We found mixed responses of using synthetically generated data for training models, with limited mitigation seen in 48% cases while remaining cases showing a boost in attack success. Table 9 shows that across all datasets, IG and SHAP saw a maximum decrease (shown in green) in attack success by 2%, SG showed < 1% decrease and LIME showed the highest decrease of 9% on models trained using synthetic data. In other cases, an increase (shown in red) in attack success was also seen by a maximum value of 2% for IG, 3% for SG, 4% for SHAP, and 27% for LIME. Hospital saw maximum increase in attack success on LIME explanations for sensitive attribute 1 (Fig. 7 (j)). Thus, both increase and decrease in privacy was observed when synthetic data was used as a privacy preservation method. Among the synthetic data generation algorithms, as seen in the last column in Table 9, CTGAN, Gaussian Copula and TVAE had similar responses, with no specific algorithm consistently achieving mitigation or boosting the attack success.

**Table 9**

Best and worst performances in pre-model across both sensitive attributes.

| XAI | Dataset | Max Decrease in Attack Success | Corresponding Decrease in Attack F1-Score | Corresponding Pre-model Method | Max Increase in Attack Success | Corresponding Increase in Attack F1-Score | Corresponding Pre-model Method |
|---|---|---|---|---|---|---|---|
| **IG** | Adult | 0.33% | 0.0025 | Gaussian Copula | 0.21% | 0.0015 | TVAE |
| | Credit | 0.45% | 0.0037 | TVAE | 0.66% | 0.007 | Gaussian Copula |
| | Compas | 2.16% | 0.0287 | TVAE | 2.33% | 0.0137 | CTGAN |
| | Hospital | 2% | 0.0218 | Gaussian Copula | 1.96% | 0.0126 | TVAE |
| **SG** | Adult | 0.54% | 0.0039 | Gaussian Copula | 0.46% | -0.0024 | Gaussian Copula |
| | Credit | 0.51% | 0.0037 | Gaussian Copula | 0.68% | 0.0067 | Gaussian Copula |
| | Compas | 0.84% | 0.0048 | CTGAN | 1.47% | 0.0176 | Gaussian Copula |
| | Hospital | 0.43% | 0.0026 | Gaussian Copula | 3.03% | 0.0334 | Gaussian Copula |
| **SHAP** | Adult | 2% | 0.0142 | TVAE | 0.5% | 0.0029 | TVAE |
| | Credit | 0.57% | 0.0066 | TVAE | 3.63% | 0.0259 | Gaussian Copula |
| | Compas | 1.69% | 0.0103 | Gaussian Copula | 0.31% | 0.0047 | Gaussian Copula |
| | Hospital | 2.31% | 0.015 | TVAE | 3.2% | 0.0354 | TVAE |
| **LIME** | Adult | 9.16% | 0.0717 | Gaussian Copula | No increase observed | NA | NA |
| | Credit | 5.43% | 0.0738 | CTGAN | 0.69% | 0.0075 | Gaussian Copula |
| | Compas | 4.93% | 0.0695 | CTGAN | 4.62% | 0.033 | TVAE |
| | Hospital | 0.29% | 0.001 | Gaussian Copula | 26.74% | 0.3346 | TVAE |

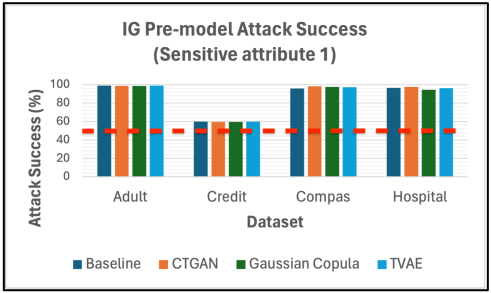

*(a)*

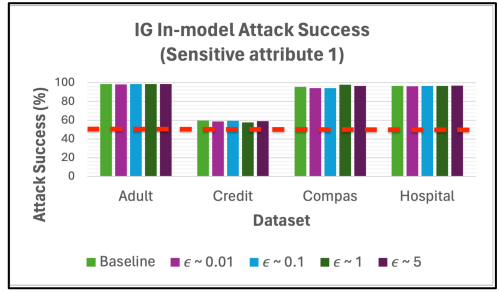

*(b)*

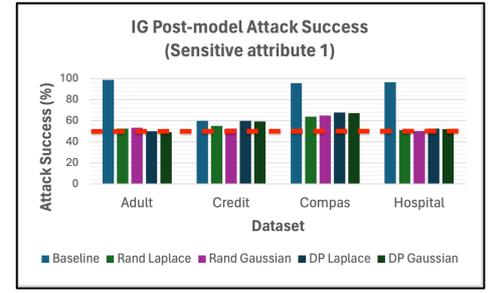

*(c)*

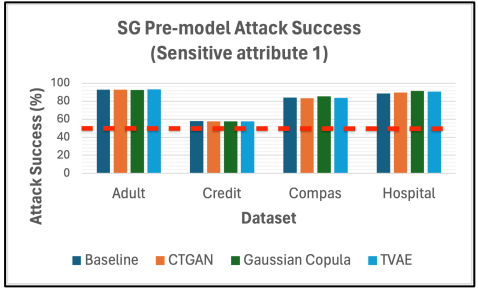

*(d)*

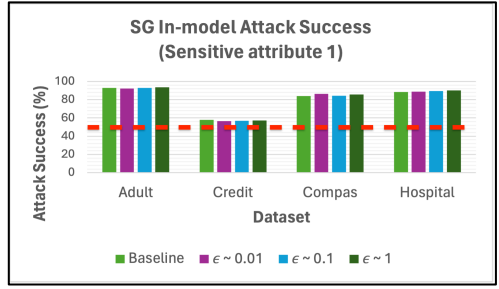

*(e)*

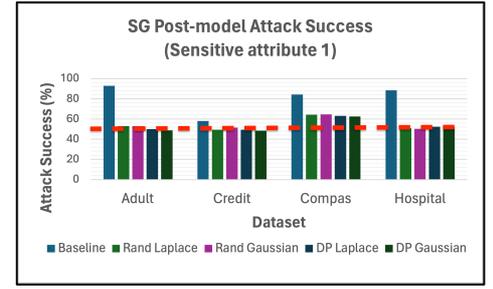

*(f)*

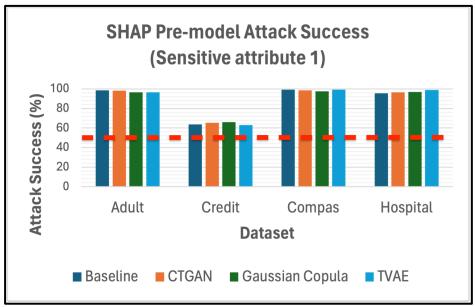

*(g)*

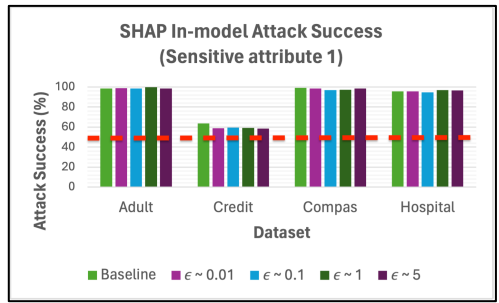

*(h)*

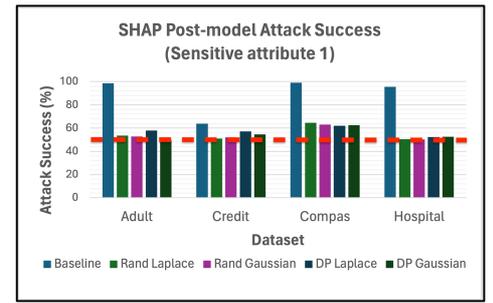

*(i)*

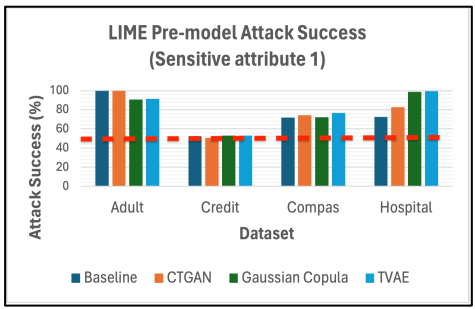

*(j)*

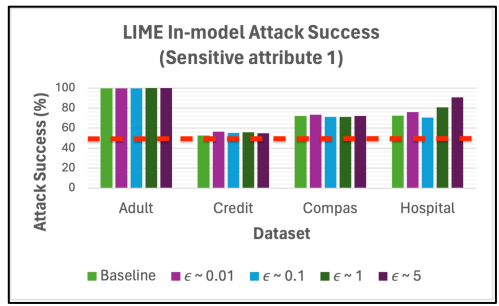

*(k)*

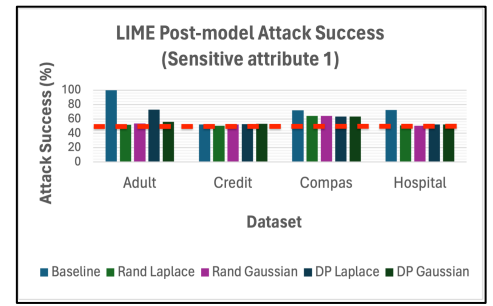

*(l)*

*Fig. 7. Pre-model, in-model and post-model attack success for different explanation types (sensitive attribute 1).*

The results from in-model in Figs. 7 and 8 indicates the ineffectiveness of DP training in mitigating attribute inference. It was observed that the tuning of $\varepsilon$ from 0.01 to 5 failed to increase privacy against the attack. Lower values of $\varepsilon$, in the strong privacy guarantee tier, which are expected to boost privacy, failed to strengthen the resilience of the system against the attack. Mixed responses were seen, with DP training helping with limited mitigation of the attack in 44% of cases and increasing the attack success in others. As seen from Table 10, the maximum decrease (shown in green) in attack success was 4% for IG and SG, 7% for SHAP and 37% for LIME. No specific tier of $\varepsilon$ was observed to have an advantage in mitigation. The maximum increase (shown in red) in attack success was 2% for IG and SHAP, 3% for SG, and 18% for LIME. Thus, similar to pre-model, the observations from in-model show both increase and decrease in privacy.

**Table 10**

Best and worst performances in in-model across both sensitive attributes.

| XAI | Dataset | Max Decrease in Attack Success | Corresponding Decrease in Attack F1-Score | Corresponding In-model Hyperparameter | Max Increase in Attack Success | Corresponding Increase in Attack F1-Score | Corresponding In-model Hyperparameter |
|---|---|---|---|---|---|---|---|
| **IG** | Adult | 0.4% | 0.0030 | $\varepsilon = 0.01$ | 0.3% | 0.0021 | $\varepsilon = 0.1$ |
| | Credit | 2.15% | 0.0227 | $\varepsilon = 1$ | 0.9% | 0.0074 | $\varepsilon = 0.1$ |
| | Compas | 4.03% | 0.0672 | $\varepsilon = 1$ | 2.19% | 0.0300 | $\varepsilon = 0.1$ |
| | Hospital | 0.82% | 0.0052 | $\varepsilon = 0.1$ | 0.88% | 0.0056 | $\varepsilon = 5$ |
| **SG** | Adult | 1.4% | 0.0099 | $\varepsilon = 5$ | 0.71% | 0.0055 | $\varepsilon = 1$ |
| | Credit | 1.42% | 0.0199 | $\varepsilon = 0.01$ | 0.54% | 0.0066 | $\varepsilon = 5$ |
| | Compas | 3.83% | 0.0119 | $\varepsilon = 1$ | 2.63% | 0.0197 | $\varepsilon = 0.1$ |
| | Hospital | 0.86% | 0.0054 | $\varepsilon = 0.1$ | 1.92% | 0.0208 | $\varepsilon = 5$ |
| **SHAP** | Adult | No drop observed | NA | NA | 1.96% | 0.0115 | $\varepsilon = 5$ |
| | Credit | 7.39% | 0.0291 | $\varepsilon = 1$ | No increase observed | NA | NA |
| | Compas | 4.93% | 0.0733 | $\varepsilon = 1$ | No increase observed | NA | NA |
| | Hospital | 0.82% | 0.0088 | $\varepsilon = 0.1$ | 2.48% | 0.0161 | $\varepsilon = 0.01$ |
| **LIME** | Adult | 0.2% | 0.0014 | $\varepsilon = 0.1$ | 2.99% | 0.0211 | $\varepsilon = 0.1$ |
| | Credit | No drop observed | NA | NA | 5.37% | 0.0594 | $\varepsilon = 0.1$ |
| | Compas | 36.91% | 0.5319 | $\varepsilon = 1$ | 1.49% | 0.0093 | $\varepsilon = 0.01$ |
| | Hospital | 2.16% | 0.0564 | $\varepsilon = 0.1$ | 18.31% | 0.2285 | $\varepsilon = 5$ |

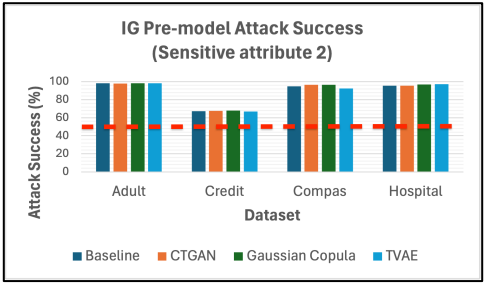
*(a)*
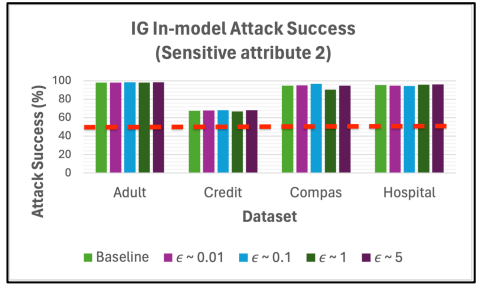
*(b)*
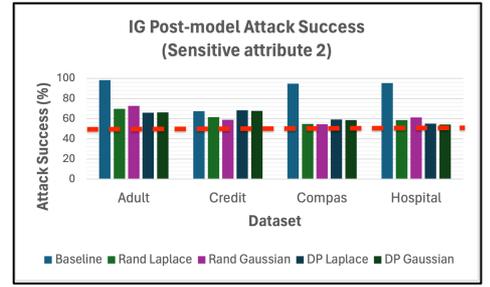
*(c)*

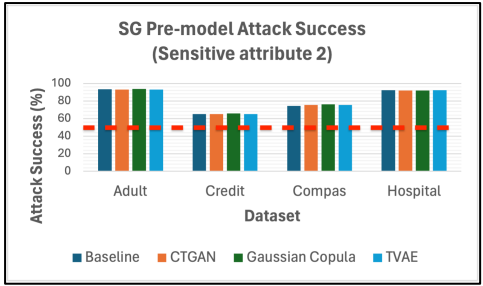
*(d)*
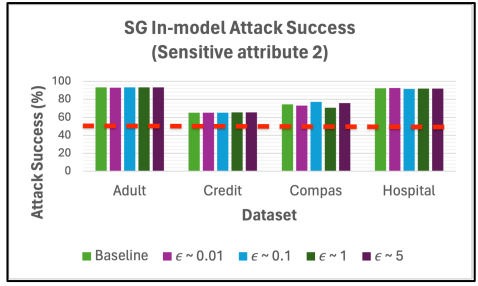
*(e)*
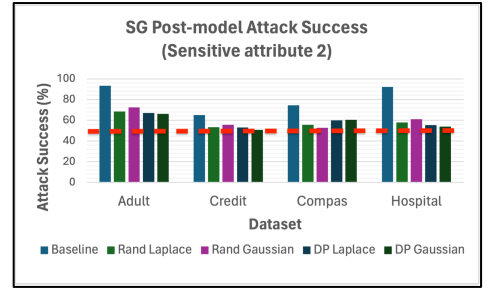
*(f)*

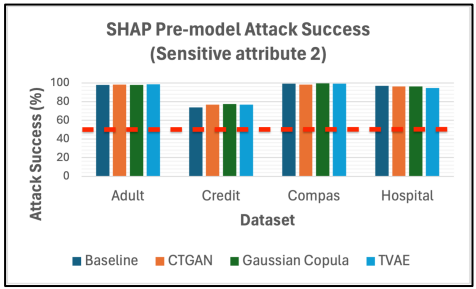
*(g)*
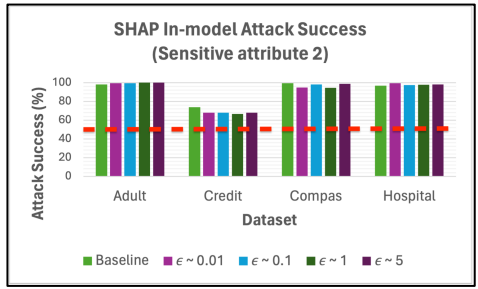
*(h)*
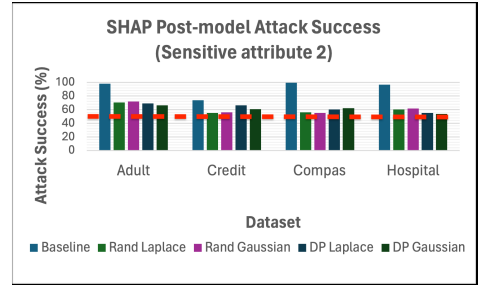
*(i)*

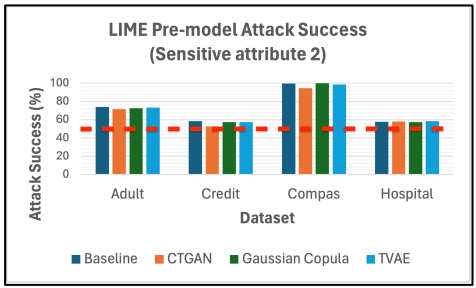
*(j)*
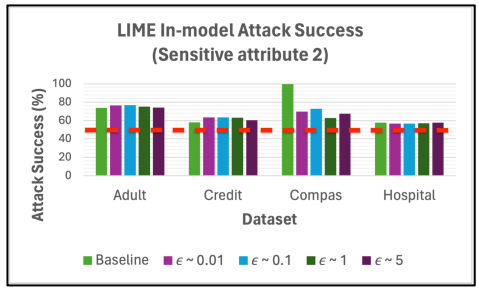
*(k)*
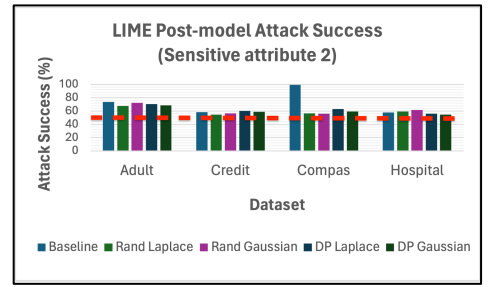
*(l)*

*Fig. 8. Pre-model, in-model and post-model attack success for different explanation types (sensitive attribute 2).*

Post-model noise methods consistently displayed reduction in attack success, bringing the attack success closer to random guess. Both random and calibrated noise worked well in mitigating the attack. From Table 11, we see that the maximum decrease (shown in green) in attack success for sensitive attribute 1 was 49% on IG and SHAP, 44% on SG and 48% on LIME. The maximum decrease (shown in green) for sensitive attribute 2 was 41% on IG, 38% on SG, and 44% on SHAP and LIME. Both Gaussian strategies, namely, random and calibrated, had higher mitigation responses across all datasets and explanation types compared to Laplace strategies. Laplace methods were found to work better with LIME. Credit being most resilient among all datasets, showed the least improvement in attack success compared to others. Among all attacks done in post-model, which consisted of total 128 attacks (equation 10), 9 cases showed a minor increase in attack success. This increase was < 4% on LIME and <1% on IG. Overall, 93% of 352 total attacks (equations 8, 9, 10) showed mitigation of the attack. Among post-model strategies, random noise performed better in 63% of cases combined in IG, SHAP and LIME and 25% of cases in SG. Calibrated noise had better performance in SG.

**Table 11**

Best performance of post-model methods.

| XAI | Dataset | Sensitive Attribute 1 | | | Sensitive Attribute 2 | | |
|---|---|---|---|---|---|---|---|
| | | Max Decrease in Attack Success | Corresponding Decrease in Attack F1-Score | Corresponding Post-model Method | Max Decrease in Attack Success | Corresponding Decrease in Attack F1-Score | Corresponding Post-model Method |
| IG | Adult | 49.47% | 0.4395 | DP Gaussian | 32.27% | 0.2050 | DP Laplace |
| | Credit | 7.32% | 0.1282 | Random Gaussian | 8.45% | 0.0937 | Random Gaussian |
| | Compas | 32.02% | 0.2116 | Random Laplace | 40.14% | 0.5896 | Random Gaussian |
| | Hospital | 46.13% | 0.5012 | Random Gaussian | 41.21% | 0.3110 | DP Gaussian |
| SG | Adult | 43.95% | 0.3992 | DP Gaussian | 27.10% | 0.1805 | DP Gaussian |
| | Credit | 9.6% | 0.1933 | DP Gaussian | 14.2% | 0.1736 | DP Gaussian |
| | Compas | 21.65% | 0.1580 | DP Gaussian | 21.93% | 0.3320 | Random Gaussian |
| | Hospital | 38.13% | 0.4199 | Random Gaussian | 38.39% | 0.2940 | DP Gaussian |
| SHAP | Adult | 48.78% | 0.4294 | DP Gaussian | 31.55% | 0.2005 | DP Gaussian |
| | Credit | 12.78% | 0.1498 | Random Laplace | 18.61% | 0.1575 | Random Laplace |
| | Compas | 37% | 0.2456 | DP Laplace | 44.39% | 0.6491 | Random Gaussian |
| | Hospital | 45.25% | 0.4884 | Random Gaussian | 43% | 0.3235 | DP Gaussian |
| LIME | Adult | 47.77% | 0.3923 | Random Laplace | 5.97% | 0.0423 | Random Laplace |

|   |   |   |   |   |   |   |
|---|---|---|---|---|---|---|
| Credit | 1.8% | 0.0408 | Random Laplace | 3.67% | 0.0467 | Random Laplace |
| Compas | 8.63% | 0.0711 | DP Laplace | 43.84% | 0.6475 | Random Gaussian |
| Hospital | 22.29% | 0.2041 | Random Gaussian | 3.31% | 0.0355 | DP Gaussian |

## 5.2. Impact on explanation quality

An important design goal of XAI systems is to produce explanations that are accurate and true to the original model [93]. Hence, to measure the alignment of explanations with the model, we use faithfulness as a measure of explanation quality. Explanations with higher faithfulness values are expected to have higher alignment with the target model, leading to more trustworthy explanations. To validate our faithfulness evaluation, we selected 3 types of metrics, namely, faithfulness correlation, faithfulness estimate and sufficiency. We measured these values before and after integration with PETs to determine any adverse effects of PETs on explanation quality.

Fig. 9 shows the results of faithfulness metrics on all datasets. We observe that faithfulness correlation and faithfulness estimate, though different in definition (Subsection 2.5), are aligned for all types of explanations except for two noted anomalies. This is seen in the form of a decrease in faithfulness estimate in Adult for random Laplace (Fig. 9 (b)) and a decrease in Credit for random Gaussian (Fig. 9 (e)). SG in Adult and Hospital shows a decrease in sufficiency for post-model methods, however, the corresponding faithfulness correlation and estimate values remain stable in both datasets. Except for these decreases, the faithfulness values remain stable around the baseline or vary slightly. Since an increase in the values of these metrics indicates higher faithfulness, we note that faithfulness of explanations generated by PETs enhanced systems are not adversely affected by privacy preservation.

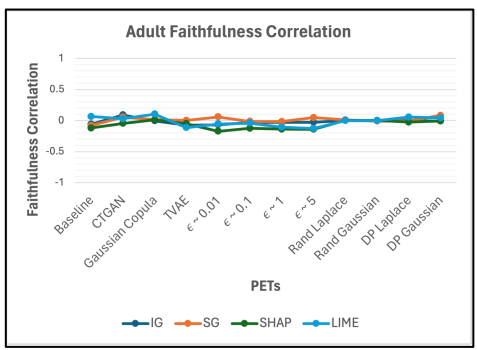
(a)

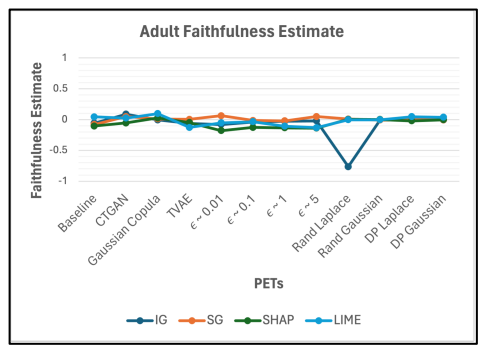
(b)

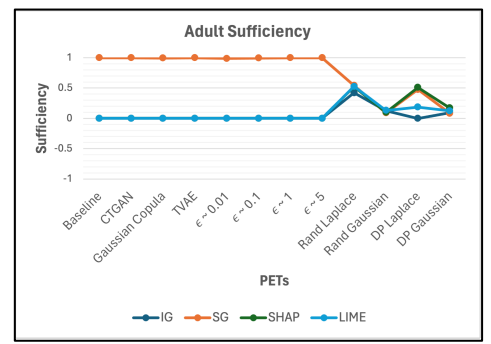
(c)

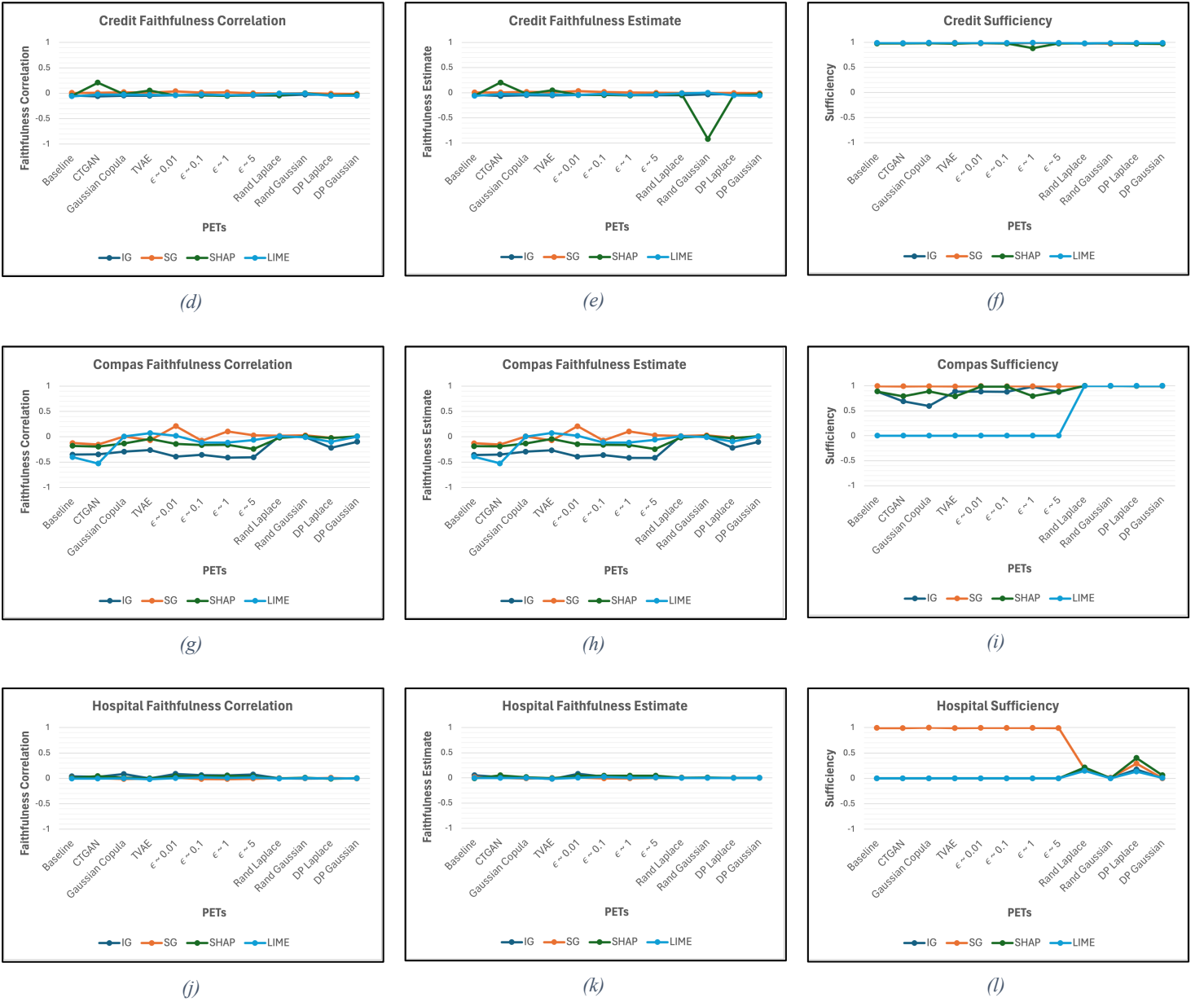

Fig. 9. XAI faithfulness for all explanations.

### 5.3. Impact on model accuracy

In this subsection, we investigate the impact of PETs on model accuracy. In pre-model setup, the models are trained using synthetic data instead of real training data. To ensure that the trained models are representative of the original problem, we test the models on the original test sets, i.e., the non-synthetic test sets. Fig. 10 shows the training and test accuracies of the synthetic models in comparison to the baseline accuracy and Table 12 shows the change in the test accuracies. All datasets, except Credit, show a decrease in test accuracy on the synthetic models. The highest decrease of 12% was seen on the model trained with synthetic version of Compas created using Gaussian Copula followed by 8% decrease on model trained on synthetic version of Adult created using CTGAN. From Table 12, we observe that TVAE algorithm shows the least and Gaussian Copula, the highest decrease in test accuracy for all datasets.

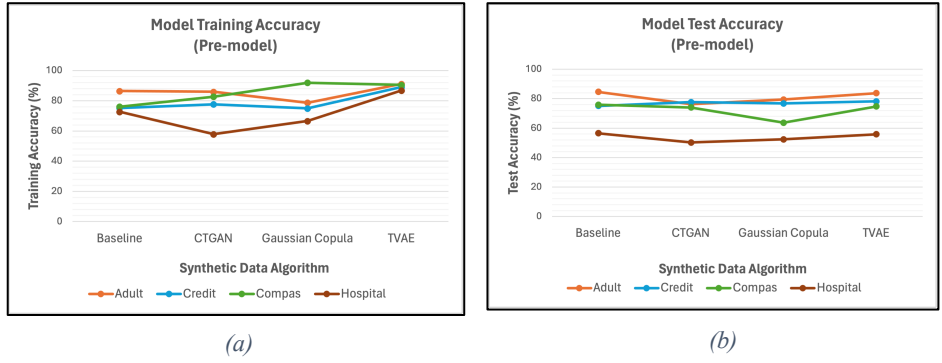

*(a)* *(b)*

Fig. 10. Training and test accuracy for baseline and synthetic models.

**Table 12**

Test accuracy change in pre-model and in-model.

| Pre-model Test Accuracy Change | | | | | In-model Test Accuracy Change | | | | |
|---|---|---|---|---|---|---|---|---|---|
| Synthetic Data Algorithm | Adult | Credit | Compas | Hospital | DP Privacy Budget | Adult | Credit | Compas | Hospital |
| CTGAN | -8.34 | 2.59 | -1.78 | -6.32 | $\varepsilon \sim 0.01$ | -10.92 | -37.20 | -12.10 | -6.59 |
| Gaussian Copula | -5.15 | 1.65 | -12.13 | -4.17 | $\varepsilon \sim 0.1$ | -8.51 | -24.34 | -12.13 | -6.06 |
| TVAE | -0.86 | 3.06 | -1.16 | -0.75 | $\varepsilon \sim 1$ | -8.43 | -11.69 | -12.49 | -6.35 |
| | | | | | $\varepsilon \sim 5$ | -8.52 | -5.73 | -12.35 | -6.06 |

In in-model setup, as reported in previous works [6], [33], [94], DP-SGD training algorithm was found to have an adverse impact on the accuracies of the trained models. As seen in Fig. 11, all datasets show a drop in accuracy, with highest decrease in the accuracy for Credit by 37% for $\varepsilon \sim 0.01$. The accuracy was found to improve with increasing values of ε for Credit and remained stable across the variations of ε for other datasets. From Table 12, we find that as reported in literature, the maximum decrease in training accuracy was observed for lower values of $\varepsilon$ and increased as the privacy budget increased.

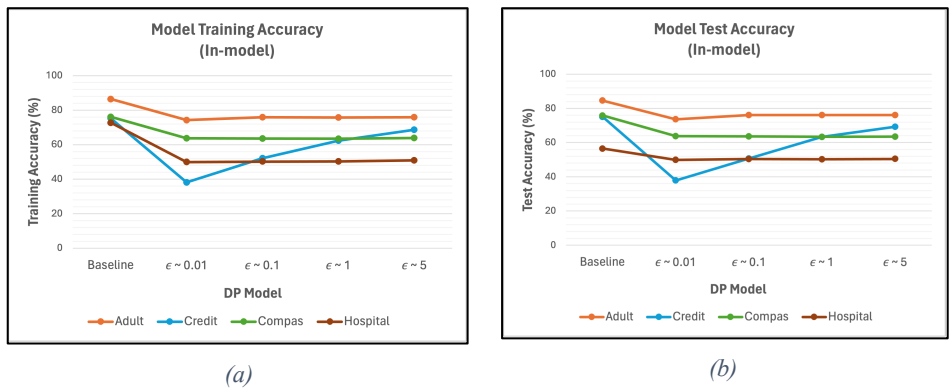

*(a)* *(b)*

Fig. 11. Training and test accuracy for baseline and DP models.

In post-model setup, retraining of models was not required. Hence there was no change in accuracy and the model utility was unaffected in this setup. Thus, overall post-model performed best in terms of accuracy, followed by pre-model which had a lower decrease in accuracy as compared to in-model.

### 5.4. Impact on performance time

In each of the 3 setups, privacy preservation methods were found to introduce certain performance overheads. In pre-model, the overhead was experienced during the process of synthetic data generation and was found to be dependent on the algorithm used. In in-model, training using DP added to the additional computational time. In post-model, the overhead was experienced during the process of adding calibrated noise to explanations. Adding random noise, however, had negligible performance overhead.

In pre-model setup, the generation of synthetic data was found to be dependent on the number of records generated and the complexity of the datasets. Table 13 summarizes the synthetic data generation time for each dataset for the 3 generation algorithms considered. As seen from the generation time/record, CTGAN was found to take the longest generation time while Gaussian Copula was the fastest. No noticeable change in the training time of the synthetic models was observed, with the training time staying within $\pm 50$ seconds of the baseline models as seen in Fig. 12 (a). For Compas and Hospital, the synthetic models trained faster than the baseline models.

**Table 13**

Performance overhead introduced by synthetic data generation in pre-model.

| Properties | Adult | Credit | Compas | Hospital |
|---|---|---|---|---|
| **Number of records** | 32,561 | 30,000 | 6,907 | 30,000 |
| **Number of columns/record** | 82 | 24 | 11 | 121 |
| **Total CTGAN time (s)** | 3,396 | 2,163.72 | 189 | 4,742.27 |
| **CTGAN time/record (ms)** | 104.3 | 72.12 | 27.36 | 158.08 |
| **Total Gaussian Copula time (secs)** | 46.27 | 13.05 | 2.15 | 65.07 |
| **Gaussian Copula time/record (ms)** | 1.42 | 0.44 | 0.31 | 2.17 |
| **Total TVAE generation time (secs)** | 1,478.24 | 745 | 100.76 | 1,407.82 |
| **TVAE time/record (ms)** | 45.40 | 24.83 | 14.59 | 46.93 |

In in-model setup, the DP models were trained on the original dataset using DP-SGD algorithm. Due to the per sample gradient clipping required by this algorithm, an increase in training time was seen across all datasets [71]. This increase in training time was also found to be dataset dependent as observed in Fig. 12 (b). The longest training times for Credit, Adult and Compas was up to 2, 11 and 19 times the original training times respectively. For Hospital, however, a decrease was observed, and the shortest DP training time was 34 secs faster than the original training time.

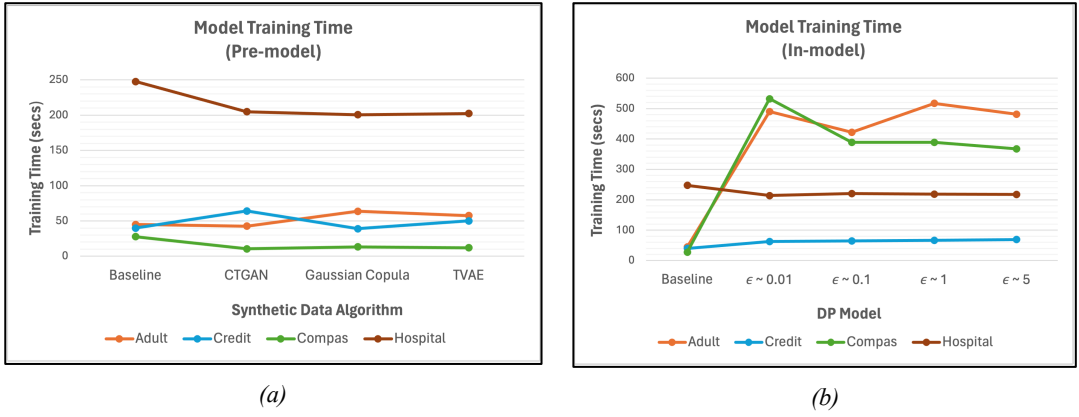

Fig. 12. Training time for pre-model and in-model stages.

Post-model setup did not see a change in the training time, but an overhead was observed when adding calibrated noise to the explanation records. This was found to be dependent on the number of columns in an explanation record and the total number of explanation records. As seen in Table 14, Hospital experienced the longest while Compas saw the shortest noise addition time. For random noise, this performance overhead was absent.

**Table 14**

Performance overhead introduced by DP noise in post-model.

| Properties | Adult | Credit | Compas | Hospital |
|---|---|---|---|---|
| **Number of explanation records** | 5373 | 4950 | 1140 | 4925 |
| **Number of columns/explanation record** | 81 | 23 | 10 | 120 |
| **Total time to add calibrated noise (s)** | 24.03 | 6.23 | 0.56 | 27.66 |
| **Total time/explanation record (ms)** | 4.47 | 1.26 | 0.49 | 5.62 |

Besides this overhead introduced by PETs, perturbation-based explanations are slow to generate due to different combinations of features that are evaluated [95]. However, since this overhead is inherent in the explanation generation algorithm and not additionally introduced by the PETs mechanism, we did not investigate this performance issue further. Other performance related issues, such as the overhead for generation of XAI metrics, especially those using perturbations, were also considered out of scope of this work.

### 5.5. Limitations

Our work is constrained within the applicability domain of the selected attribute inference attack used for privacy auditing. The attack is independent of the target model architecture and assumes the adversary possesses an auxiliary dataset with black-box access to the model. The attack is applicable to tabular datasets and used for inference of binary sensitive attributes. Our experiments are conducted on a limited number of feature-based XAI, i.e., two XAI methods each from backpropagation and perturbation-based categories. We measured explanation quality using faithfulness and other XAI metrics such as complexity and robustness, are not evaluated.

## 6. Discussion of Results

In our empirical evaluation, we use 3 groups of metrics for privacy, explainability and utility respectively. In this section, we discuss the main results for each measured property and make recommendations for privacy preservation of feature-based XAI.

### 6.1. Privacy Preservation

The evaluation in our study uses attack effectiveness to measure privacy. The scores from attack success and attack F1-score metrics were found to be in alignment and hence we base our discussions on the attack success scores. In the absence of PETs, among backpropagation-based methods, SG was found to be more resilient to attribute inference than IG. This may stem from the use of raw gradients by IG which reveal more information about the underlying attributes. Among perturbation-based methods, LIME was observed to be more resilient than SHAP. This could arise from local explanations that LIME produces around the point of interest which could not be accurately learnt by the attack model to reverse an attribute.

In pre-model stage of the AI pipeline, limited mitigation was seen in 48% of the attacks while 52% showed an increase in attack effectiveness. The results seen with different synthetic data generation algorithms, namely, CTGAN, Gaussian Copula and TVAE, were similar, with no algorithm showing an advantage in mitigation compared to others. Thus models trained on synthetic data are susceptible to attribute inference in XAI as determined in other works on non-XAI systems [96], [97].

In in-model stage of the AI pipeline, limited mitigation was seen in 44% of attacks while 56% of attacks showed an improvement. Thus, mitigation in in-model was weaker than in pre-model. The use of different values of privacy budgets in DP training was ineffective in improving privacy against attribute inference. Lower tiers of privacy budgets failed to reduce the attack effectiveness. This can be due to the mechanics of DP, which effectively hides the presence of a single record in a dataset, making it more suitable for mitigating membership inference [6] than attribute inference [56].

Among the 3 stages of the AI pipeline, post-model was found to be effective with consistent improvement in attack mitigation in 93% of attacks. This can be attributed to the use of noise in explanations to introduce adversarial examples to mislead the attack classifier [75]. Both random and calibrated noise brought the attack success closer to random guess. The effect of random and calibrated noise were similar, with random working better in IG, SHAP and LIME, and calibrated working better in SG. Gaussian noise techniques were observed to reduce attack success more than Laplace techniques, except in LIME in which Laplace noise worked better.

### 6.2. Explanation Quality

Out of the 3 different faithfulness metrics used for evaluating explanation quality, the observations from faithfulness correlation were found to be aligned with faithfulness estimate. Hence, we note that these two metrics can be interchangeably used. The observations on faithfulness correlation, faithfulness estimate, and sufficiency were found to be stable around baseline or vary slightly across all setups. Thus, introducing PETs did not adversely affect the faithfulness and hence the quality of explanations.

## 6.3. Model Utility

We used model accuracy to measure the utility of the model. A drop in accuracy was observed when synthetic data was used to train the models. Among the synthetic data generation algorithms, Gaussian Copula had the highest while TVAE had the lowest drop in accuracy. Similar drop in accuracy was seen when DP training was used in in-model. Among the various values of privacy budget used, lower values had higher decrease in accuracy. The drop in accuracy in in-model was higher than that observed in pre-model. In post-model, the model accuracy was unaltered since the original target models were used for explanation generation.

## 6.4. Performance Time

Each setup was found to introduce a performance overhead due to privacy preservation measures. In pre-model, the overhead was introduced due to the additional time required for generation of synthetic datasets. Among the synthetic data generation algorithms, Gaussian Copula was observed to have the shortest while CTGAN the longest generation time/record as determined in a previous study [98]. In in-model, training using DP-SGD algorithm introduced an additional training time which was highest for Compas and up to 19 times its original training time. In post-model, calibrated noise introduced a minor performance overhead in comparison to random noise that had negligible cost.

Among the 3 setups, pre-model introduced the highest performance overhead in terms of runtime for CTGAN and TVAE algorithms. This was higher than the additional time required for DP training in in-model. Post-model methods were the fastest in spite of the overhead of generating calibrated noise. Thus overall, post-model techniques improved privacy while retaining explanation faithfulness, model utility and introducing least additional computational time.

## 6.5. Recommendations

From the main results deduced from this study of comparison of PETs in XAI, we have consolidated the following recommendations for usage of XAI and integration of PETs to maximize their benefits with privacy, explainability and performance:

- Without the use of any PETs with XAI, SG is more resilient than IG in backpropagation-based methods while LIME is more resilient than SHAP in perturbation-based methods. Hence these inherently resilient methods may be selected when training or query data consists of sensitive attributes.
- When used as PETs in XAI, synthetic data in pre-model and differentially private training in in-model, do not mitigate attribute inference and hence are not recommended for privacy preservation against this attack.
- Noise addition in post-model preserves privacy from attribute inference in comparison to synthetic data and differentially private training. Noise techniques reduce attack effectiveness towards the level of random guess.
- Noise addition in post-model achieves privacy preservation in explanations while maintaining model accuracy, explanation faithfulness and introducing least performance overhead.
- Gaussian noise, both random and calibrated, lower attack success more effectively than Laplace noise in IG, SG and SHAP. Laplace noise techniques work better with LIME.
- Random noise introduces negligible performance overhead and produces similar results to calibrated noise. Random noise can be used effectively for privacy preservation in IG, SHAP and LIME, while calibrated is effective in SG.

- If synthetic data is used as PETs, among the 3 synthetic data generation algorithms, Gaussian Copula is the fastest at the cost of accuracy of the trained model while TVAE gives reasonable generation time along with accuracy.
- Multiple PETs can be implemented in an XAI system due to their complimentary nature. The use of hybrid privacy preservation approaches can benefit the privacy of the overall system. However, some PETs may adversely affect other system properties. For instance, differential privacy decreases the accuracy of the model at low privacy budgets and in general increases the training time. Hence the impact of PETs on other system properties should be evaluated to maximize the benefits of hybrid approaches.

## 7. Conclusion

XAI methods are important for AI transparency but suffer from privacy leakage of training and query data. To counter this issue and build privacy resilience, it is intuitive to consider the use of PETs in conjunction with these methods. In this article, we have empirically evaluated the use of synthetic data, differential privacy, and noise addition in different stages of the AI pipeline to boost the privacy of the system and mitigate a known privacy attack that inferences sensitive attributes from training and query data. We measured privacy, explanation faithfulness, model utility and system performance to determine methods that can achieve a balance of these various properties. Our findings determine that some stages in the AI pipeline work better than others in balancing these properties in an XAI system. Our evaluation suggests that post-model techniques are stronger in introducing privacy preservation in feature-based XAI while maintaining the system utility in terms of model accuracy and explanation faithfulness, and introducing least performance cost. Thus, our results indicate that it is possible to introduce privacy preservation without compromising the utility of the system.

## 8. References

[1] M. Nassar, K. Salah, M. H. ur Rehman, and D. Svetinovic, "Blockchain for explainable and trustworthy artificial intelligence," *WIREs Data Min. Knowl. Discov.*, vol. 10, no. 1, Jan. 2020, doi: 10.1002/widm.1340.

[2] G. Yang, Q. Ye, and J. Xia, "Unbox the black-box for the medical explainable AI via multi-modal and multi-centre data fusion: A mini-review, two showcases and beyond," *Inf. Fusion*, vol. 77, pp. 29–52, Jan. 2022, doi: 10.1016/j.inffus.2021.07.016.

[3] R. Shokri, M. Strobel, and Y. Zick, "On the Privacy Risks of Model Explanations," in *Proceedings of the 2021 AAAI/ACM Conference on AI, Ethics, and Society*, Virtual Event USA: ACM, Jul. 2021, pp. 231–241. doi: 10.1145/3461702.3462533.

[4] F. Naretto, A. Monreale, and F. Giannotti, "Evaluating the Privacy Exposure of Interpretable Global Explainers," presented at the 2022 IEEE 4th International Conference on Cognitive Machine Intelligence (CogMI), Atlanta, GA, USA: IEEE Computer Society, 2022, pp. 13–19. doi: 10.1109/CogMI56440.2022.00012.

[5] M. Pawelczyk, H. Lakkaraju, and S. Neel, "On the Privacy Risks of Algorithmic Recourse," in *Proceedings of The 26th International Conference on Artificial Intelligence and Statistics*, in Proceedings of Machine Learning Research, vol. 206. Valencia, Spain: PMLR, 2023, pp. 9680–9696. [Online]. Available: https://proceedings.mlr.press/v206/pawelczyk23a.html

[6] H. Liu, Y. Wu, Z. Yu, and N. Zhang, "Please Tell Me More: Privacy Impact of Explainability through the Lens of Membership Inference Attack," presented at the IEEE Symposium on Security and Privacy (SP), San Francisco, CA, USA, 2024, pp. 119–119. doi: 10.1109/SP54263.2024.00120.

[7] A. Kuppa and N.-A. Le-Khac, "Adversarial XAI Methods in Cybersecurity," *IEEE Trans. Inf. Forensics Secur.*, vol. 16, pp. 4924–4938, 2021, doi: 10.1109/TIFS.2021.3117075.

[8] V. Duddu and A. Boutet, "Inferring Sensitive Attributes from Model Explanations," in *Proceedings of the 31st ACM International Conference on Information & Knowledge Management (CIKM '22)*, New York, NY, USA: Association for Computing Machinery, Oct. 2022, pp. 416–425. doi: 10.1145/3511808.3557362.

[9] X. Luo, Y. Jiang, and X. Xiao, "Feature Inference Attack on Shapley Values," in *Proceedings of the 2022 ACM SIGSAC Conference on Computer and Communications Security*, Los Angeles CA USA: ACM, Nov. 2022, pp. 2233–2247. doi: 10.1145/3548606.3560573.

[10] X. Zhao, W. Zhang, X. Xiao, and B. Lim, "Exploiting Explanations for Model Inversion Attacks," in *2021 IEEE/CVF International Conference on Computer Vision (ICCV)*, Montreal, QC, Canada: IEEE, Oct. 2021, pp. 662–672. doi: 10.1109/ICCV48922.2021.00072.

[11] S. Goethals, K. Sörensen, and D. Martens, "The Privacy Issue of Counterfactual Explanations: Explanation Linkage Attacks," *ACM Trans. Intell. Syst. Technol.*, vol. 14, no. 5, pp. 1–24, Oct. 2023, doi: 10.1145/3608482.

[12] S. Milli, L. Schmidt, A. D. Dragan, and M. Hardt, "Model Reconstruction from Model Explanations," in *Proceedings of the Conference on Fairness, Accountability, and Transparency*, Atlanta GA USA: ACM, Jan. 2019, pp. 1–9. doi: 10.1145/3287560.3287562.

[13] U. Aïvodji, A. Bolot, and S. Gambs, "Model extraction from counterfactual explanations," Sep. 03, 2020, *arXiv*: arXiv:2009.01884. Accessed: Feb. 24, 2023. [Online]. Available: http://arxiv.org/abs/2009.01884

[14] Y. Wang, H. Qian, and C. Miao, "DualCF: Efficient Model Extraction Attack from Counterfactual Explanations," in *2022 ACM Conference on Fairness, Accountability, and Transparency*, Seoul Republic of Korea: ACM, Jun. 2022, pp. 1318–1329. doi: 10.1145/3531146.3533188.

[15] A. Yan, T. Huang, L. Ke, X. Liu, Q. Chen, and C. Dong, "Explanation leaks: Explanation-guided model extraction attacks," *Inf. Sci.*, vol. 632, pp. 269–284, Jun. 2023, doi: 10.1016/j.ins.2023.03.020.

[16] A. Yan, R. Hou, H. Yan, and X. Liu, "Explanation-based data-free model extraction attacks," *World Wide Web*, vol. 26, no. 5, pp. 3081–3092, Sep. 2023, doi: 10.1007/s11280-023-01150-6.

[17] A. Blanco-Justicia, J. Domingo-Ferrer, S. Martínez, and D. Sánchez, "Machine learning explainability via microaggregation and shallow decision trees," *Knowl.-Based Syst.*, vol. 194, p. 105532, Apr. 2020, doi: 10.1016/j.knosys.2020.105532.

[18] J. Zhao, H. Zhu, F. Wang, R. Lu, and H. Li, "Efficient and privacy-preserving tree-based inference via additive homomorphic encryption," *Inf. Sci.*, vol. 650, p. 119480, Dec. 2023, doi: 10.1016/j.ins.2023.119480.

[19] A. Goldsteen, G. Ezov, and A. Farkash, "Reducing Risk of Model Inversion Using Privacy-Guided Training," Jun. 29, 2020, *arXiv*: arXiv:2006.15877. Accessed: Apr. 05, 2023. [Online]. Available: http://arxiv.org/abs/2006.15877

[20] S. Yeom, I. Giacomelli, M. Fredrikson, and S. Jha, "Privacy Risk in Machine Learning: Analyzing the Connection to Overfitting," in *2018 IEEE 31st Computer Security Foundations Symposium (CSF)*, Oxford: IEEE, Jul. 2018, pp. 268–282. doi: 10.1109/CSF.2018.00027.

[21] R. Shokri, M. Strobel, and Y. Zick, "Exploiting Transparency Measures for Membership Inference: a Cautionary Tale," in *The AAAI Workshop on Privacy-Preserving Artificial Intelligence (PPAI)*, New York, USA: AAAI, 2020.

[22] N. Patel, R. Shokri, and Y. Zick, "Model Explanations with Differential Privacy," in *2022 ACM Conference on Fairness, Accountability, and Transparency*, Seoul Republic of Korea: ACM, Jun. 2022, pp. 1895–1904. doi: 10.1145/3531146.3533235.

[23] G. Cohen and R. Giryes, "Membership Inference Attack Using Self Influence Functions," in *2024 IEEE/CVF Winter Conference on Applications of Computer Vision (WACV)*, Waikoloa, HI, USA: IEEE, Jan. 2024, pp. 4880–4889. doi: 10.1109/WACV57701.2024.00482.

[24] H. Montenegro, W. Silva, and J. S. Cardoso, "Privacy-Preserving Generative Adversarial Network for Case-Based Explainability in Medical Image Analysis," *IEEE Access*, vol. 9, pp. 148037–148047, 2021, doi: 10.1109/ACCESS.2021.3124844.

[25] H. Montenegro and J. S. Cardoso, "Anonymizing medical case-based explanations through disentanglement," *Med. Image Anal.*, vol. 95, p. 103209, Jul. 2024, doi: 10.1016/j.media.2024.103209.

[26] F. Campos, L. Petrychenko, L. F. Teixeira, and W. Silva, "Latent Diffusion Models for Privacy-preserving Medical Case-based Explanations," in *1st Workshop on Explainable Artificial Intelligence for the Medical Domain, EXPLIMED 2024*, Santiago de Compostela; Spain: CEUR-WS, 2024.

[27] V. Vo *et al.*, "Feature-based Learning for Diverse and Privacy-Preserving Counterfactual Explanations," in *Proceedings of the 29th ACM SIGKDD Conference on Knowledge Discovery and Data Mining*, Long Beach CA USA: ACM, Aug. 2023, pp. 2211–2222. doi: 10.1145/3580305.3599343.

[28] E. Johnson, S. Mohan, A. Gaudio, A. Smailagic, C. Faloutsos, and A. Campilho, "HeartSpot: Privatized and Explainable Data Compression for Cardiomegaly Detection," in *2022 IEEE-EMBS International Conference on Biomedical and Health Informatics (BHI)*, Ioannina, Greece: IEEE, Sep. 2022, pp. 01–04. doi: 10.1109/BHI56158.2022.9926777.

[29] A. Gaudio *et al.*, "DeepFixCX: Explainable privacy-preserving image compression for medical image analysis," *WIREs Data Min. Knowl. Discov.*, Mar. 2023, doi: 10.1002/widm.1495.

[30] H. Jeong, S. Lee, S. J. Hwang, and S. Son, "Learning to Generate Inversion-Resistant Model Explanations," presented at the 36th Conference on Neural Information Processing Systems (NeurIPS 2022), New Orleans, 2022.

[31] Y. Wu, S. Cai, X. Xiao, G. Chen, and B. C. Ooi, "Privacy preserving vertical federated learning for tree-based models," *Proc. VLDB Endow.*, vol. 13, no. 12, pp. 2090–2103, Aug. 2020, doi: 10.14778/3407790.3407811.

[32] M. Molhoek and J. V. Laanen, "Secure Counterfactual Explanations in a Two-party Setting," in *2024 27th International Conference on Information Fusion (FUSION)*, Venice, Italy: IEEE, Jul. 2024, pp. 1–10. doi: 10.23919/FUSION59988.2024.10706413.

[33] A. Blanco-Justicia, D. Sánchez, J. Domingo-Ferrer, and K. Muralidhar, "A Critical Review on the Use (and Misuse) of Differential Privacy in Machine Learning," *ACM Comput. Surv.*, vol. 55, no. 8, pp. 1–16, Aug. 2023, doi: 10.1145/3547139.

[34] F. Harder, M. Bauer, and M. Park, "Interpretable and Differentially Private Predictions," *Proc. AAAI Conf. Artif. Intell.*, vol. 34, no. 04, pp. 4083–4090, Apr. 2020, doi: 10.1609/aaai.v34i04.5827.

[35] S. Berning, V. Dunning, D. Spagnuelo, T. Veugen, and J. Van Der Waa, "The Trade-off Between Privacy & Quality for Counterfactual Explanations," in *Proceedings of the 19th International Conference on Availability, Reliability and Security*, Vienna Austria: ACM, Jul. 2024, pp. 1–9. doi: 10.1145/3664476.3670897.

[36] A. B. Arrieta *et al.*, "Explainable Artificial Intelligence (XAI): Concepts, Taxonomies, Opportunities and Challenges toward Responsible AI," *Inf. Fusion*, vol. 58, pp. 82–115, 2020, doi: https://doi.org/10.1016/j.inffus.2019.12.012.

[37] E. Rajabi and K. Etminani, "Knowledge-graph-based explainable AI: A systematic review," *J. Inf. Sci.*, p. 016555152211128, Sep. 2022, doi: 10.1177/01655515221112844.

[38] R. Dwivedi *et al.*, "Explainable AI (XAI): Core Ideas, Techniques, and Solutions," *ACM Comput. Surv.*, vol. 55, no. 9, pp. 1–33, Sep. 2023, doi: 10.1145/3561048.

[39] M. Ancona, E. Ceolini, C. Öztireli, and M. Gross, "Towards better understanding of gradient-based attribution methods for Deep Neural Networks," in *6th International Conference on Learning Representations, ICLR 2018*, Vancouver, BC, Canada: OpenReview.net, Apr. 2018. doi: 10.3929/ethz-b-000249929.

[40] M. Sundararajan, A. Taly, and Q. Yan, "Axiomatic Attribution for Deep Networks," in *Proceedings of the 34th International Conference on Machine Learning - Volume 70*, in ICML'17. Sydney, NSW, Australia: JMLR.org, 2017, pp. 3319–3328. doi: 10.5555/3305890.3306024.

[41] D. Smilkov, N. Thorat, B. Kim, F. Viégas, and M. Wattenberg, "SmoothGrad: removing noise by adding noise," in *Proceedings of the International Conference on Machine Learning—Workshop on Visualization for Deep Learning*, Sydney, Australia: ICML, Aug. 2017.

[42] S. M. Lundberg and S.-I. Lee, "A Unified Approach to Interpreting Model Predictions," in *Proceedings of the 31st International Conference on Neural Information Processing Systems*, in NIPS'17. Long Beach, California, USA: Curran Associates Inc., Dec. 2017, pp. 4768–4777. doi: 10.5555/3295222.3295230.

[43] C. Molnar, *Interpretable Machine Learning*. 2023.

[44] M. T. Ribeiro, S. Singh, and C. Guestrin, "'Why Should I Trust You?': Explaining the Predictions of Any Classifier," in *Proceedings of the 22nd ACM SIGKDD International Conference on Knowledge Discovery and Data Mining*, San Francisco California USA: ACM, Aug. 2016, pp. 1135–1144. doi: 10.1145/2939672.2939778.

[45] R. Shokri, M. Stronati, C. Song, and V. Shmatikov, "Membership Inference Attacks Against Machine Learning Models," in *2017 IEEE Symposium on Security and Privacy (SP)*, San Jose, CA, USA: IEEE, May 2017, pp. 3–18. doi: 10.1109/SP.2017.41.

[46] M. Fredrikson, S. Jha, and T. Ristenpart, "Model Inversion Attacks that Exploit Confidence Information and Basic Countermeasures," in *Proceedings of the 22nd ACM SIGSAC Conference on Computer and Communications Security*, Denver Colorado USA: ACM, Oct. 2015, pp. 1322–1333. doi: 10.1145/2810103.2813677.

[47] Y. Ma, X. Zhai, D. Yu, Y. Yang, X. Wei, and Y. Chen, "Label-Only Membership Inference Attack Based on Model Explanation," *Neural Process. Lett.*, vol. 56, no. 5, p. 236, Sep. 2024, doi: 10.1007/s11063-024-11682-1.

34[48] J. Ferry, U. Aïvodji, S. Gambs, M.-J. Huguet, and M. Siala, "Probabilistic Dataset Reconstruction from Interpretable Models," in *2024 IEEE Conference on Secure and Trustworthy Machine Learning (SaTML)*, Toronto, ON, Canada: IEEE, Apr. 2024, pp. 1–17. doi: 10.1109/SaTML59370.2024.00009.

[49] R. Toma and H. Kikuchi, "Combinations of AI Models and XAI Metrics Vulnerable to Record Reconstruction Risk," in *Privacy in Statistical Databases*, vol. 14915, J. Domingo-Ferrer and M. Önen, Eds., in Lecture Notes in Computer Science, vol. 14915. , Cham: Springer Nature Switzerland, 2024, pp. 329–343. doi: 10.1007/978-3-031-69651-0_22.

[50] T. Miura, T. Shibahara, and N. Yanai, "MEGEX: Data-Free Model Extraction Attack Against Gradient-Based Explainable AI," in *Proceedings of the 2nd ACM Workshop on Secure and Trustworthy Deep Learning Systems*, Singapore Singapore: ACM, Jul. 2024, pp. 56–66. doi: 10.1145/3665451.3665533.

[51] A. Yan, R. Hou, X. Liu, H. Yan, T. Huang, and X. Wang, "Towards explainable model extraction attacks," *Int. J. Intell. Syst.*, vol. 37, no. 11, pp. 9936–9956, Nov. 2022, doi: 10.1002/int.23022.

[52] J. Ye, A. Maddi, S. K. Murakonda, V. Bindschaedler, and R. Shokri, "Enhanced Membership Inference Attacks against Machine Learning Models," in *Proceedings of the 2022 ACM SIGSAC Conference on Computer and Communications Security*, in CCS '22. Los Angeles, CA, USA: Association for Computing Machinery, Sep. 2022, pp. 3093–3106. doi: 10.1145/3548606.3560675.

[53] N. Ponomareva *et al.*, "How to DP-fy ML: A Practical Guide to Machine Learning with Differential Privacy," *J. Artif. Intell. Res.*, vol. 77, pp. 1113–1201, Jul. 2023, doi: 10.1613/jair.1.14649.

[54] M. Slokom, P.-P. De Wolf, and M. Larson, "Exploring Privacy-Preserving Techniques on Synthetic Data as a Defense Against Model Inversion Attacks," in *Information Security*, vol. 14411, E. Athanasopoulos and B. Mennink, Eds., in Lecture Notes in Computer Science, vol. 14411. , Cham: Springer Nature Switzerland, 2023, pp. 3–23. doi: 10.1007/978-3-031-49187-0_1.

[55] C.-L. Chen, L. Golubchik, and R. Pal, "Achieving Transparency Report Privacy in Linear Time," *J. Data Inf. Qual.*, vol. 14, no. 2, pp. 1–56, Jun. 2022, doi: 10.1145/3460001.

[56] Y. Zhang, R. Jia, H. Pei, W. Wang, B. Li, and D. Song, "The Secret Revealer: Generative Model-Inversion Attacks Against Deep Neural Networks," in *2020 IEEE/CVF Conference on Computer Vision and Pattern Recognition (CVPR)*, Seattle, WA, USA: IEEE, Jun. 2020, pp. 250–258. doi: 10.1109/CVPR42600.2020.00033.

[57] T. Wang, Y. Zhang, and R. Jia, "Improving Robustness to Model Inversion Attacks via Mutual Information Regularization," *Proc. AAAI Conf. Artif. Intell.*, vol. 35, no. 13, pp. 11666–11673, May 2021, doi: 10.1609/aaai.v35i13.17387.

[58] T. A. O. Alves, F. M. G. França, and S. Kundu, "MLPrivacyGuard: Defeating Confidence Information based Model Inversion Attacks on Machine Learning Systems," in *Proceedings of the 2019 Great Lakes Symposium on VLSI*, Tysons Corner VA USA: ACM, May 2019, pp. 411–415. doi: 10.1145/3299874.3319457.

[59] J. Jia and N. Z. Gong, "AttriGuard: A Practical Defense Against Attribute Inference Attacks via Adversarial Machine Learning".

[60] S. Z. El Mestari, G. Lenzini, and H. Demirci, "Preserving data privacy in machine learning systems," *Comput. Secur.*, vol. 137, p. 103605, Feb. 2024, doi: 10.1016/j.cose.2023.103605.

[61] A.-T. Tran, T.-D. Luong, and V.-N. Huynh, "A comprehensive survey and taxonomy on privacy-preserving deep learning," *Neurocomputing*, vol. 576, p. 127345, Apr. 2024, doi: 10.1016/j.neucom.2024.127345.

[62] X. Yin, Y. Zhu, and J. Hu, "A Comprehensive Survey of Privacy-preserving Federated Learning: A Taxonomy, Review, and Future Directions," *ACM Comput. Surv.*, vol. 54, no. 6, pp. 1–36, Jul. 2022, doi: 10.1145/3460427.

[63] A. Guerra-Manzanares, L. J. L. Lopez, M. Maniatakos, and F. E. Shamout, "Privacy-Preserving Machine Learning for Healthcare: Open Challenges and Future Perspectives," in *Trustworthy Machine Learning for Healthcare*, vol. 13932, H. Chen and L. Luo, Eds., in Lecture Notes in Computer Science, vol. 13932. , Cham: Springer Nature Switzerland, 2023, pp. 25–40. doi: 10.1007/978-3-031-39539-0_3.

[64] R. R. Palle and K. C. R. Kathala, "Privacy-Preserving AI Techniques," in *Privacy in the Age of Innovation*, Berkeley, CA: Apress, 2024, pp. 47–61. doi: 10.1007/979-8-8688-0461-8_5.

[65] J.-F. Rajotte, R. Bergen, D. L. Buckeridge, K. El Emam, R. Ng, and E. Strome, "Synthetic data as an enabler for machine learning applications in medicine," *iScience*, vol. 25, no. 11, p. 105331, Nov. 2022, doi: 10.1016/j.isci.2022.105331.

[66] W. Abbasi, P. Mori, and A. Saracino, "Further Insights: Balancing Privacy, Explainability, and Utility in Machine Learning-based Tabular Data Analysis," in *Proceedings of the 19th International Conference on Availability, Reliability and Security*, Vienna Austria: ACM, Jul. 2024, pp. 1–10. doi: 10.1145/3664476.3670901.

[67] L. Xu, M. Skoularidou, A. Cuesta-Infante, and K. Veeramachaneni, "Modeling Tabular data using Conditional GAN," in *Advances in Neural Information Processing Systems*, Curran Associates, Inc., 2019. [Online]. Available: https://proceedings.neurips.cc/paper_files/paper/2019/file/254ed7d2de3b23ab10936522dd547b78-Paper.pdf

[68] N. Patki, R. Wedge, and K. Veeramachaneni, "The Synthetic Data Vault," in *2016 IEEE International Conference on Data Science and Advanced Analytics (DSAA)*, Montreal, QC, Canada: IEEE, Oct. 2016, pp. 399–410. doi: 10.1109/DSAA.2016.49.

[69] C. Dwork, "Differential Privacy," in *Automata, Languages and Programming*, M. Bugliesi, B. Preneel, V. Sassone, and I. Wegener, Eds., Berlin, Heidelberg: Springer Berlin Heidelberg, 2006, pp. 1–12.

[70] C. Dwork, K. Kenthapadi, F. McSherry, I. Mironov, and M. Naor, "Our Data, Ourselves: Privacy Via Distributed Noise Generation," in *Advances in Cryptology - EUROCRYPT 2006*, vol. 4004, S. Vaudenay, Ed., in Lecture Notes in Computer Science, vol. 4004. , Berlin, Heidelberg: Springer Berlin Heidelberg, 2006, pp. 486–503. doi: 10.1007/11761679_29.

[71] M. Abadi *et al.*, "Deep Learning with Differential Privacy," in *Proceedings of the 2016 ACM SIGSAC Conference on Computer and Communications Security*, Vienna Austria: ACM, Oct. 2016, pp. 308–318. doi: 10.1145/2976749.2978318.

[72] Z. Islam and L. Brankovic, "Noise Addition for Protecting Privacy in Data Mining," presented at the Engineering Mathematics and Applications Conference, Engineering Mathematics Group, ANZIAM, 2003, pp. 85–90.

[73] J. Domingo-Ferrer, F. Sebé, and J. Castellà-Roca, "On the Security of Noise Addition for Privacy in Statistical Databases," in *Privacy in Statistical Databases*, vol. 3050, J. Domingo-Ferrer and V. Torra, Eds., in Lecture Notes in Computer Science, vol. 3050. , Berlin, Heidelberg: Springer Berlin Heidelberg, 2004, pp. 149–161. doi: 10.1007/978-3-540-25955-8_12.

[74] J. J. Kim and W. E. Winkler, "Multiplicative Noise for Masking Continuous Data," Statistical Research Division, U.S. Bureau of the Census, Washington D.C., Statistics #2003-01, Apr. 2003.

[75] J. Jia, A. Salem, M. Backes, Y. Zhang, and N. Z. Gong, "MemGuard: Defending against Black-Box Membership Inference Attacks via Adversarial Examples," in *Proceedings of the 2019 ACM SIGSAC Conference on Computer and Communications Security*, London United Kingdom: ACM, Nov. 2019, pp. 259–274. doi: 10.1145/3319535.3363201.

[76] H. Hu, Z. Salcic, L. Sun, G. Dobbie, P. S. Yu, and X. Zhang, "Membership Inference Attacks on Machine Learning: A Survey," *ACM Comput. Surv.*, vol. 54, no. 11s, pp. 1–37, Jan. 2022, doi: 10.1145/3523273.

[77] A. Hedström *et al.*, "Quantus: An Explainable AI Toolkit for Responsible Evaluation of Neural Network Explanations and Beyond," *J. Mach. Learn. Res.*, vol. 24, no. 34, pp. 1–11, 2023.

[78] U. Bhatt, A. Weller, and J. M. F. Moura, "Evaluating and Aggregating Feature-based Model Explanations," in *Proceedings of the Twenty-Ninth International Joint Conference on Artificial Intelligence*, Yokohama, Japan: International Joint Conferences on Artificial Intelligence Organization, Jul. 2020, pp. 3016–3022. doi: 10.24963/ijcai.2020/417.

[79] D. A. Melis and T. Jaakkola, "Towards Robust Interpretability with Self-Explaining Neural Networks," *Adv. Neural Inf. Process. Syst.*, vol. 31, 2018.

[80] S. Dasgupta, N. Frost, and M. Moshkovitz, "Framework for Evaluating Faithfulness of Local Explanations," in *International Conference on Machine Learning*, PMLR, Jun. 2022, pp. 4794–4815.

[81] B. Becker and R. Kohavi, "Adult." UCI Machine Learning Repository, 1996. doi: https://doi.org/10.24432/C5XW20.

[82] I.-C. Yeh, "Default of Credit Card Clients." UCI Machine Learning Repository, 2016. doi: https://doi.org/10.24432/C55S3H.

[83] Propublica, "Compas." 2016. doi: https://www.propublica.org/datastore/dataset/compas-recidivism-risk-score-data-and-analysis.

[84] P. W. Koh and P. Liang, "Understanding Black-box Predictions via Influence Functions," in *Proceedings of Machine Learning Research*, Jul. 2017, pp. 1885–1894. [Online]. Available: https://proceedings.mlr.press/v70/koh17a

[85] B. Strack *et al.*, "Impact of HbA1c Measurement on Hospital Readmission Rates: Analysis of 70,000 Clinical Database Patient Records," *BioMed Res. Int.*, vol. 2014, pp. 1–11, 2014, doi: 10.1155/2014/781670.

[86] A. Maximilian, M. Haegele, P. Seegerer, K. Schuett, A. Hill, and S. Lapuschkin, "INNvestigate," INNvestigate Neural Networks. Accessed: May 02, 2024. [Online]. Available: https://github.com/albermax/innvestigate

[87] S. Lundberg, "SHAP," Shapley Additive Explanations. Accessed: May 02, 2024. [Online]. Available: https://github.com/shap/shap

[88] M. Ribeiro, "LIME," Local Interpretable Model-Agnostic Explanations. Accessed: May 02, 2024. [Online]. Available: https://github.com/marcotcr/lime

[89] DataCebo, Inc., "The Synthetic Data Vault," The Synthetic Data Vault. [Online]. Available: https://sdv.dev/

[90] DataCebo, Inc., "Synthetic Data Vault Diagnostic," Synthetic Data Vault Diagnostic. [Online]. Available: https://docs.sdv.dev/sdv/single-table-data/evaluation/diagnostic

[91] C. Dwork and A. Roth, "The algorithmic foundations of differential privacy," *Found. Trends Theor. Comput. Sci.*, vol. 9, no. Nos. 3–4, pp. 211–407, 2014, doi: 10.1561/0400000042.

[92] *Google Differential Privacy Library*. Google. [Online]. Available: https://github.com/google/differential-privacy

[93] Z. Atf and P. R. Lewis, "Human Centricity in the Relationship Between Explainability and Trust in AI," *IEEE Technol. Soc. Mag.*, vol. 42, no. 4, pp. 66–76, Dec. 2023, doi: 10.1109/MTS.2023.3340238.

[94] A. Rahman, T. Rahman, R. Laganiere, N. Mohammed, and Y. Wang, "Membership Inference Attack against Differentially Private Deep Learning Model," *Trans Data Priv*, vol. 11, no. 1, pp. 61–79, 2018.

[95] A. Kapishnikov, T. Bolukbasi, F. Viegas, and M. Terry, "XRAI: Better Attributions Through Regions," in *2019 IEEE/CVF International Conference on Computer Vision (ICCV)*, Seoul, Korea (South): IEEE, Oct. 2019, pp. 4947–4956. doi: 10.1109/ICCV.2019.00505.

[96] M. S. M. S. Annamalai, A. Gadotti, and L. Rocher, "A Linear Reconstruction Approach for Attribute Inference Attacks against Synthetic Data," in *33rd USENIX Security Symposium (USENIX Security 24)*, Philadelphia, PA: USENIX Association, Aug. 2024, pp. 2351–2368. [Online]. Available: https://www.usenix.org/conference/usenixsecurity24/presentation/annamalai-linear

[97] S. Kwatra and V. Torra, "Empirical Evaluation of Synthetic Data Created by Generative Models via Attribute Inference Attack," in *Privacy and Identity Management. Sharing in a Digital World*, vol. 695, F. Bieker, S. De Conca, N. Gruschka, M. Jensen, and I. Schiering, Eds., in IFIP Advances in Information and Communication Technology, vol. 695. , Cham: Springer Nature Switzerland, 2024, pp. 282–291. doi: 10.1007/978-3-031-57978-3_18.

[98] M. Endres, A. Mannarapotta Venugopal, and T. S. Tran, "Synthetic Data Generation: A Comparative Study," in *International Database Engineered Applications Symposium*, Budapest Hungary: ACM, Aug. 2022, pp. 94–102. doi: 10.1145/3548785.3548793.

# Appendix

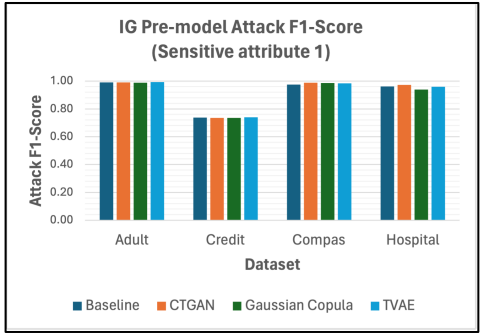

*(a)*

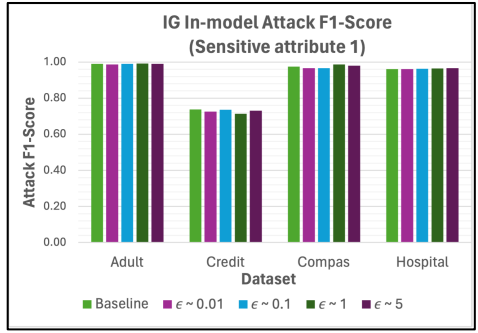

*(b)*

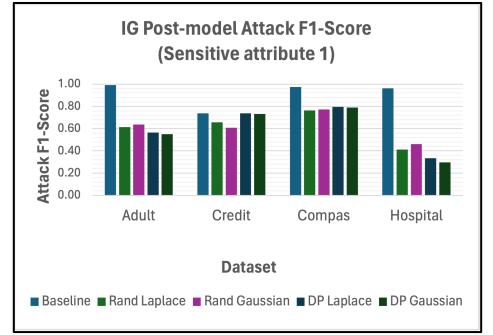

*(c)*

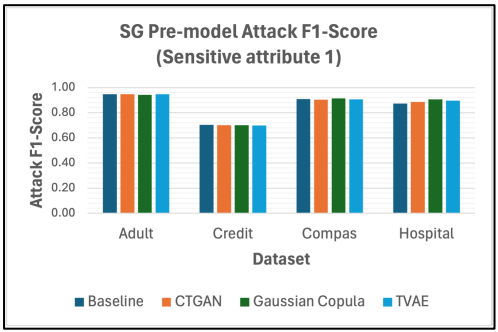

*(d)*

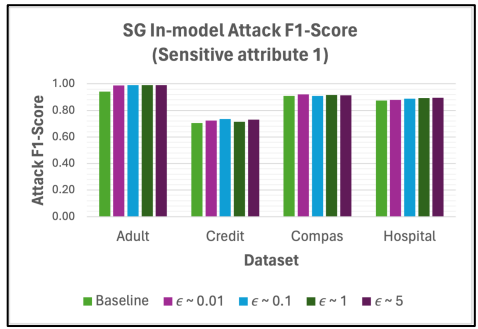

*(e)*

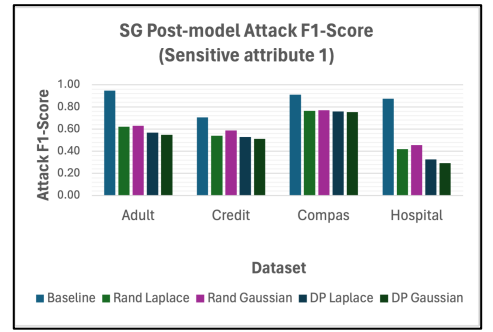

*(f)*

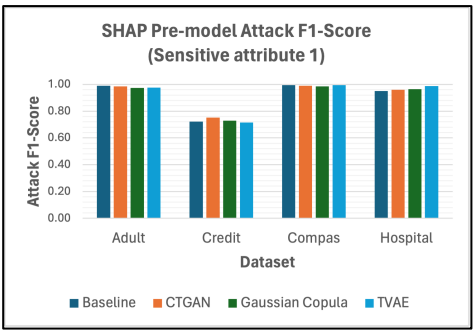

*(g)*

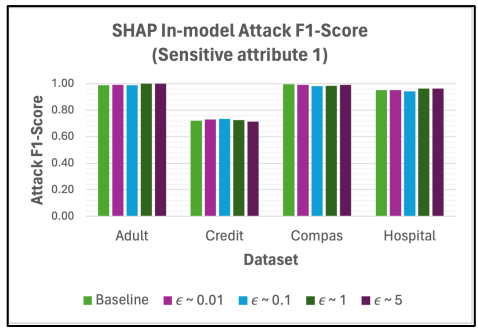

*(h)*

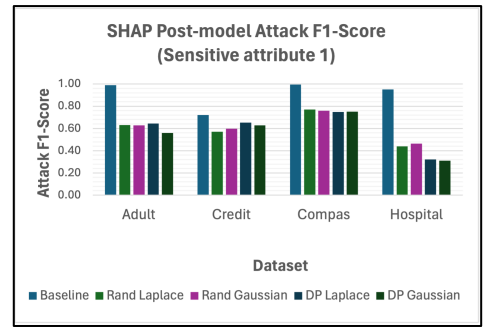

*(i)*

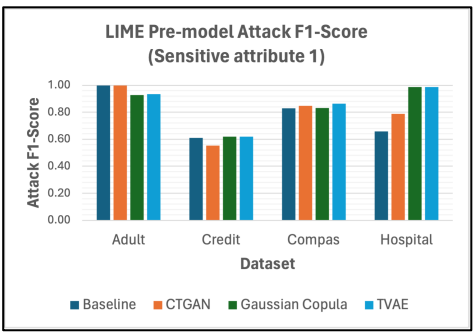

*(j)*

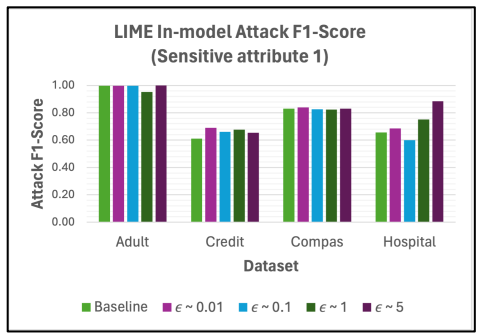

*(k)*

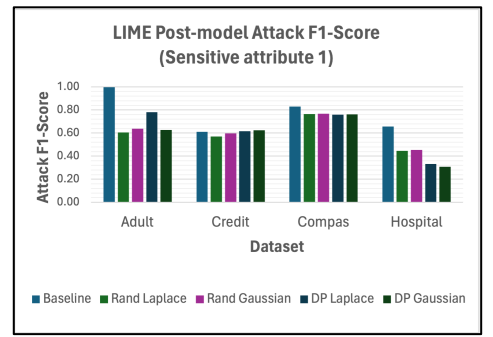

*(l)*

*Fig. 13. Pre-model, in-model and post-model attack F1-scores for all explanations (sensitive attribute 1).*

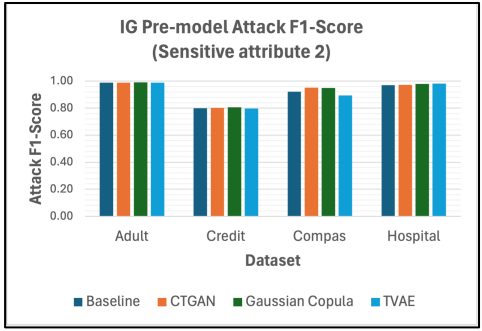
*(a)*
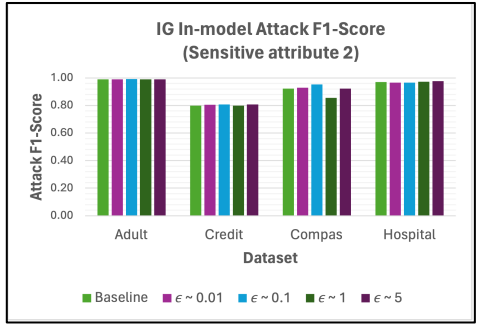
*(b)*
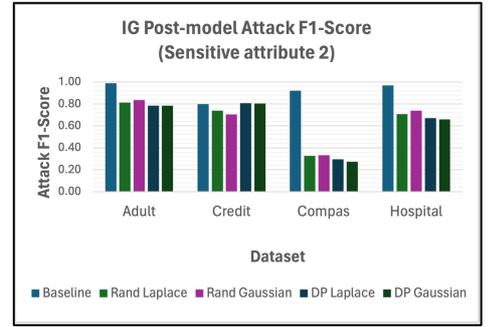
*(c)*
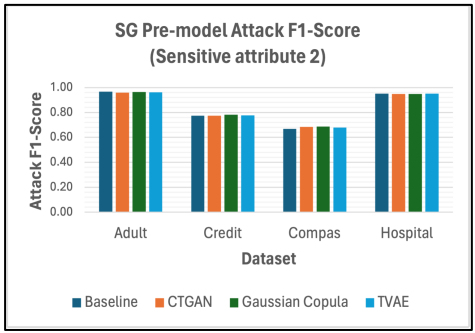
*(d)*
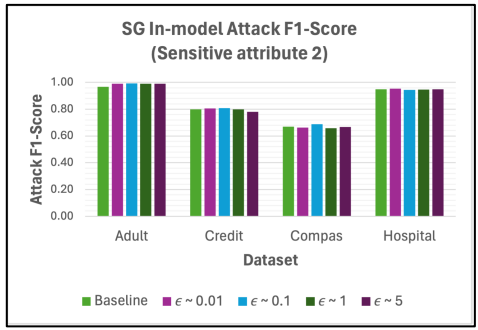
*(e)*
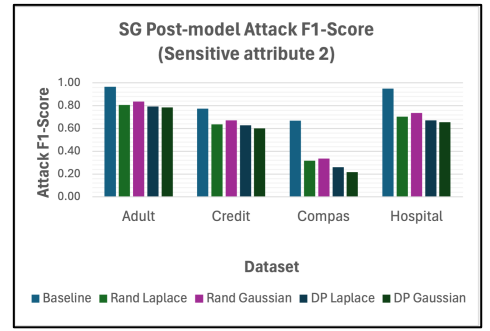
*(f)*
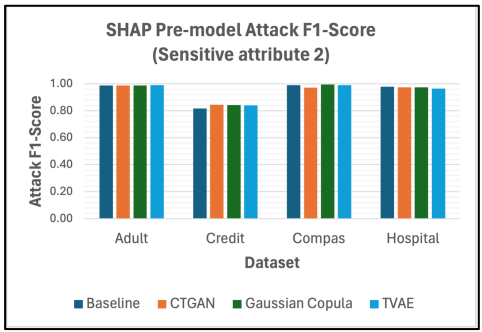
*(g)*
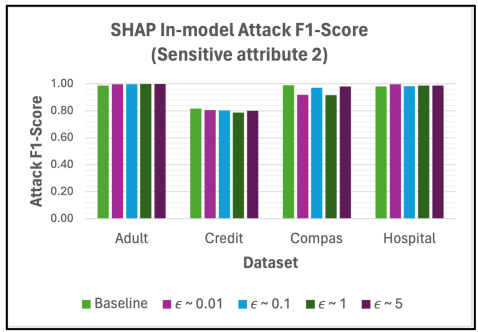
*(h)*
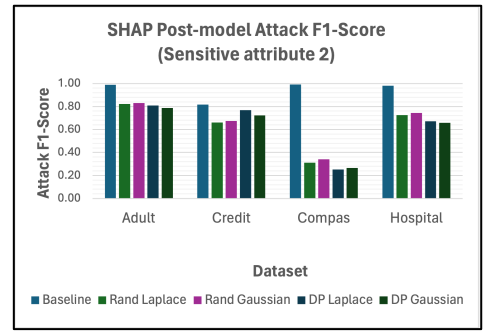
*(i)*
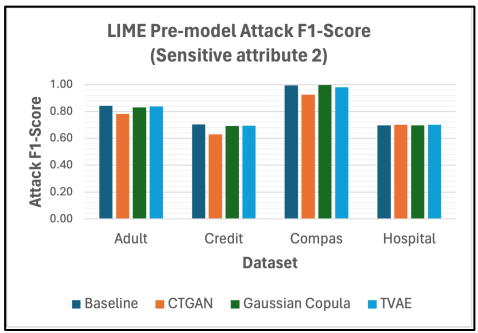
*(j)*
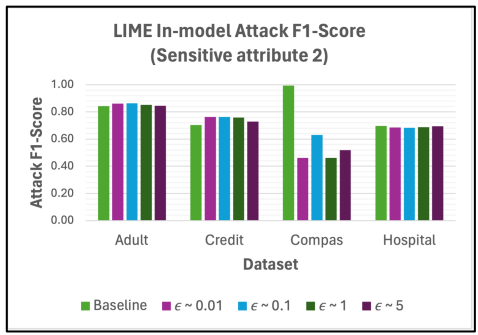
*(k)*
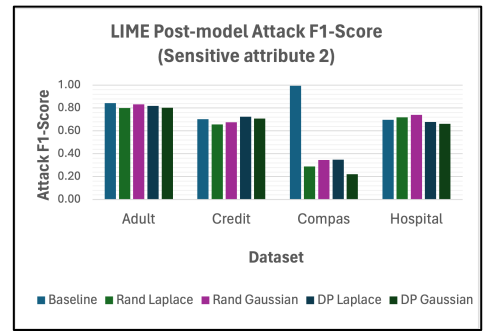
*(l)*

Fig. 14. Pre-model, in-model and post-model attack F1-scores for all explanations (sensitive attribute 2).